\documentclass[10pt,twocolumn,letterpaper]{article}

\usepackage[pagenumbers]{configs/cvpr} 

\usepackage{color}
\usepackage{amsthm}
\usepackage{wrapfig}
\usepackage{array}
\usepackage{newlfont} %
\usepackage{textcomp}
\usepackage{multirow}
\usepackage{hhline}
\usepackage{tabularx}
\usepackage{bigstrut}
\usepackage{afterpage}
\usepackage{algorithmic}
\usepackage{microtype}
\usepackage{mathtools}
\usepackage{booktabs} %
\usepackage[ruled]{algorithm2e} %
\usepackage{dingbat}

\usepackage{comment}
\usepackage{times}
\usepackage{epsfig}
\usepackage{graphicx}
\usepackage{amsmath}
\usepackage{amssymb}
\usepackage{cuted}
\usepackage{capt-of}
\usepackage{float}
\usepackage{enumitem}
\usepackage{balance}
\usepackage{bbding} 
\usepackage[normalem]{ulem} 

\usepackage[dvipsnames]{xcolor}
\definecolor{mypink3}{cmyk}{0, 0.7808, 0.4429, 0.1412}
\definecolor{mygray}{gray}{0.6}

\usepackage[toc,page]{appendix}
\usepackage[numbers,sort,compress]{natbib}

\usepackage{gensymb}

\usepackage[pagebackref,breaklinks,colorlinks]{hyperref}

\usepackage{color, colortbl}

\usepackage{amssymb}
\usepackage{pifont}

\newcommand{\Tref}[1]{Tab.~\ref{#1}}

\newcommand{\tref}[1]{Tab.~\ref{#1}}
\newcommand{\eref}[1]{Eq.~(\ref{#1})}
\newcommand{\fref}[1]{Fig.~\ref{#1}}

\newcommand{\TODO}[1]{\textcolor{red}{#1}}

\newcommand{\ourtitle}{Human-Aware Object Placement for Visual Environment Reconstruction}

\newcommand{\modelname}{\textcolor{black}{MOVER}\xspace}
\newcommand{\modelnameLong}{\textcolor{black}{``human Motion driven Object placement for Visual Environment Reconstruction''}\xspace}

\definecolor{DeltaColor}{rgb}{0.039,0.73,0.71}
\definecolor{SigmaColor}{rgb}{0.98,0.45,0.0}
\definecolor{AlphaColor}{rgb}{0,0,0.8}
\definecolor{BetaColor}{rgb}{0.8,0,0.8}
\definecolor{GammaColor}{rgb}{0.514,0.34,0.224}
\definecolor{EpsilonColor}{rgb}{0.353,0.725,0.906}
\definecolor{PurpleColor}{rgb}{0.5,0,0.7}
\definecolor{OrangeColor}{rgb}{0.914,0.541,0.0.141}
\definecolor{GreenColor}{rgb}{0.137,0.573,0.565}
\definecolor{RedColor}{rgb}{0.949,0.275, 0.224}
\definecolor{LightCyan}{rgb}{0.88,1,1}
\definecolor{Gray}{gray}{0.85}

\definecolor{BetaColor}{rgb}{0.8,0,0.8}

\newcommand{\JT}[1]{\textcolor{OrangeColor}{[JT: #1]}}
\definecolor{LightCyan}{rgb}{0.88,1,1}
\definecolor{lightgray}{rgb}{0.9,0.9,0.9}

\renewcommand{\etc}{etc\xspace}
\renewcommand{\etal}{et al.\xspace}
\renewcommand{\ie}{i.e.\xspace}
\renewcommand{\eg}{e.g.\xspace}

\newcommand{\twoD}{2D\xspace}
\newcommand{\threeD}{{3D}\xspace}

\newcommand{\rgb}{RGB\xspace}
\newcommand{\cad}{CAD\xspace}

\newcommand{\hoi}{HOI\xspace}
\newcommand{\hps}{\mbox{HPS}\xspace}

\newcommand{\hsi}{\mbox{HSI}\xspace}
\newcommand{\HSI}{\hsi}

\newcommand{\openpose}{\mbox{OpenPose}\xspace}
\newcommand{\smpl}{\mbox{SMPL}\xspace}
\newcommand{\smplx}{\mbox{SMPL-X}\xspace}
\newcommand{\smplifyx}{\mbox{SMPLify-X}\xspace}

\newcommand{\prox}{\mbox{PROX}\xspace}
\newcommand{\posa}{\mbox{POSA}\xspace}

\newcommand{\pare}{\mbox{PARE}\xspace}
\newcommand{\pigraphs}{{\mbox{PiGraphs}}\xspace}
\newcommand{\iMapper}{{\mbox{iMapper}}\xspace}
\newcommand{\phosa}{{\mbox{PHOSA}}\xspace}
\newcommand{\holisticplus}{{\mbox{Holistic++}}\xspace}
\newcommand{\holisticmesh}{{\mbox{HolisticMesh}}\xspace}

\newcommand{\stateoftheart}{{\mbox{state-of-the-art}}\xspace}

\newcommand{\supmat}{\mbox{see more details in Sup.~Mat.}\xspace}
\newcommand{\Supmat}{\mbox{See more details in Sup.~Mat.}\xspace}

\newcommand{\qheading}[1]{\noindent\textbf{#1}}
\newcommand{\tabincell}[2]{\begin{tabular}{@{}#1@{}}#2\end{tabular}} 

\newcommand{\cmark}{\ding{51}}%
\newcommand{\xmark}{\ding{55}}%
\definecolor{GreenColor}{rgb}{0.137,0.573,0.565}

\newcommand{\colorRef}[1]{\textcolor{black}{#1}} 
\usepackage[capitalize]{cleveref}
\crefname{figure}{\colorRef{Fig.}}{\colorRef{Figs.}}
\Crefname{figure}{\colorRef{Fig.}}{\colorRef{Figs.}}

\crefname{section}{\colorRef{Sec.}}{\colorRef{Secs.}}
\Crefname{section}{\colorRef{Sec.}}{\colorRef{Secs.}}
\Crefname{table}{\colorRef{Tab.}}{\colorRef{Tabs.}}
\crefname{table}{\colorRef{Tab.}}{\colorRef{Tabs.}}

\newcommand{\projectURL}{\url{https://mover.is.tue.mpg.de}}

\renewcommand{\paragraph}[1]{\medskip\noindent\textbf{#1}}

\def\cvprPaperID{923}
\def\confName{CVPR}
\def\confYear{2022}

\begin{document}

\title{\ourtitle}

\author{Hongwei Yi$^{1}$\quad Chun-Hao P. Huang$^1$\quad Dimitrios Tzionas$^{1}$\quad Muhammed Kocabas$^{1,2}$ \\
 Mohamed Hassan$^{1}$\quad Siyu Tang$^{2}$\quad Justus Thies$^{1}$\quad Michael J. Black$^{1}$ \\
$^{1}$Max Planck Institute for Intelligent Systems, T\"ubingen, Germany ~~~ $^2$ ETH Z\"urich\\ 
{\tt\small \{firstname.lastname\}@\{tuebingen.mpg.de,inf.ethz.ch\} } 
}


\newcommand{\teaserCaption}{
From a monocular video sequence, \modelname reconstructs a \threeD scene that best affords humans interacting with it. 
Existing methods for monocular \threeD scene reconstruction ignore people and produce physically implausible scenes.
\modelname takes as input:
(1)     several images of human-scene interaction (\hsi) from a static camera, 
(2)     a rough estimate of \threeD object shape and placement in \threeD space \cite{nie2020total3dunderstanding}, and
(3)     estimated \threeD human bodies interacting with the scene \cite{pavlakos2019expressive,Kocabas_PARE_2021}.
%
Each frame contains valuable information about humans, objects, and the proximal relationship between them.
\modelname accumulates this information across frames, 
to optimize for a physically plausible \threeD scene. 
The final \threeD scene is more accurate than the input and enables reasoning about human-scene contact.
}

\twocolumn[{
    \renewcommand\twocolumn[1][]{#1}
    \maketitle
    \centering
    \vspace{-1.0 em}
    \begin{minipage}{1.00\textwidth}
        \centering
        \includegraphics[trim=000mm 000mm 000mm 000mm, clip=False, width=\linewidth]{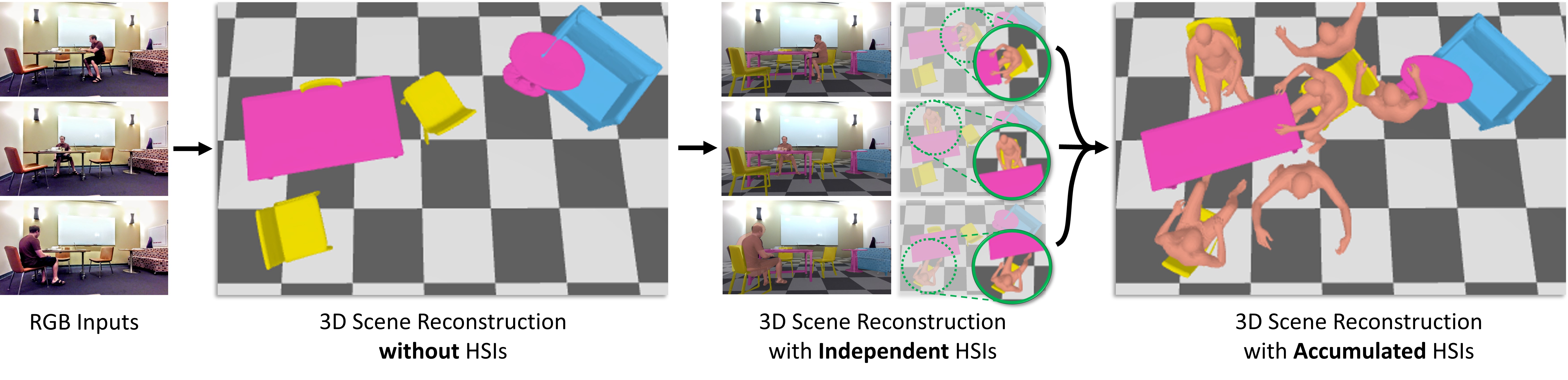} 
    \end{minipage}
    \vspace{-1.0 em}
    \captionof{figure}{\teaserCaption}
    \label{fig:teaser}
    \vspace*{+1.85 em}
}]

\begin{abstract}
\vspace{-1.0 em}
Humans are in constant contact with the world as they move through it and interact with it.
This contact is a vital source of information for understanding \threeD humans, \threeD scenes, and the interactions between them.
In fact, we demonstrate that these human-scene interactions (\textsl{\hsi}s) can be leveraged to improve the \threeD reconstruction of a scene from a monocular \rgb video.
Our key idea is that, as a person moves through a scene and interacts with it, we accumulate \textsl{\hsi}s across multiple input images, and use these in optimizing the 3D scene to reconstruct a consistent, physically plausible, \threeD scene layout.
Our optimization-based approach exploits three types of \hsi constraints:
(1) humans who move in a scene are occluded by, or occlude, objects, thus constraining the depth ordering of the objects,
(2) humans move through free space and do not interpenetrate objects, 
(3) when humans and objects are in contact, the contact surfaces occupy the same place in space.
Using these constraints in an optimization formulation across all observations, we significantly improve \threeD scene layout reconstruction.
Furthermore, we show that our scene reconstruction can be used to refine the initial \threeD human pose and shape (\hps) estimation.
We evaluate the \threeD scene layout reconstruction and \hps estimates qualitatively and quantitatively using the \prox and \pigraphs datasets. 
The code and data are available for research purposes at \projectURL.
\end{abstract}

\section{Introduction}
Human behavior and the interaction of humans with their environment are fundamentally about the \threeD world.
Hence, \threeD reconstruction of both the human and scene can facilitate 
behavior analysis.
Where and how humans interact with a scene can be used to predict future motions and interactions for human-centered AI and robots, or to synthesize these for AR/VR and other computer-graphics applications.
Tremendous progress has been made in reconstructing \threeD human bodies~\cite{Kocabas_PARE_2021, joo2020eft, Kanazawa2018_hmr, kocabas2020vibe, SPIN:ICCV:2019, pavlakos2019expressive, Pavlakos_2019_ICCV, yuan2022glamr, yuan2021simpoe, Luo2021DynamicsRegulatedKP,BEV,SHAPY_2022} 
and \threeD scenes \cite{huang2018cooperative, nie2020total3dunderstanding, Zhang_2021_CVPR, dahnert2021panoptic, bozic2021transformerfusion} from monocular images or videos, typically in isolation from each other. 
In real life, though, humans always interact with scenes. 
Consequently, humans (partially) occlude the scene, and the scene (partially) occludes humans. 
Strong human-scene occlusion can cause problems for both scene and human reconstruction.
In contrast, recent work on human-scene interaction (\HSI), estimates humans and scenes together \cite{hassan2019resolving, chen2019holistic++, weng2020holistic}.
\prox~\cite{hassan2019resolving} demonstrates how \HSI can be used to constrain \threeD human pose estimation, but it requires a \threeD scan of the full scene to be known a priori.
This is often unrealistic and cumbersome, as it requires one to conduct offline \threeD reconstruction by walking around the scene with a depth sensor \cite{zollhofer2018sotaRconstructionRGBD} to observe it from many view points.

\begin{figure}
    \centering
    \subfloat[3D scene reconstruction \cite{nie2020total3dunderstanding} and HPS \cite{pavlakos2019expressive} in isolation.]{\includegraphics[width=1.0\columnwidth]{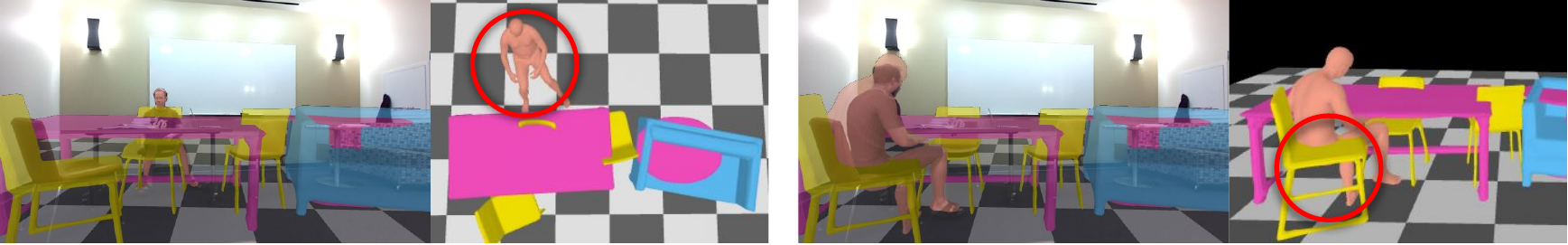}}
    \\
    \subfloat[HolisticMesh \cite{weng2020holistic}. L: single-image results. R: multiple-images result.]{\includegraphics[width=1.0\columnwidth]{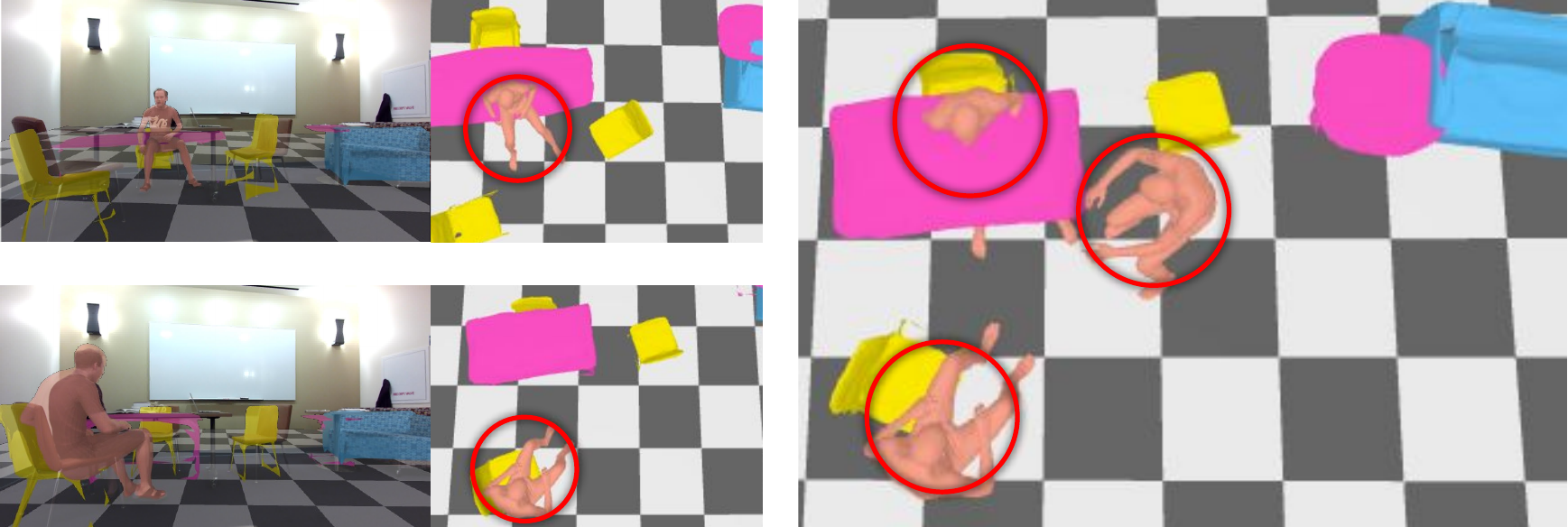}}
    \caption{                   
                                Where existing 
                                methods struggle: (a) humans in estimated scenes penetrate objects or lack contact 
                                with objects and ``hover'' in the air when estimated in isolation~\cite{nie2020total3dunderstanding,pavlakos2019expressive} %
                                (b) humans interpenetrate objects, even, when the \threeD scenes and humans are jointly optimized with single (left) or sequential images (right)~\cite{weng2020holistic}.
                                In contrast, we leverage human scene interaction constraints in a global optimization across all input frames, to compute a scene that is coherent with the human motions (see \Cref{fig:teaser}).
}
    \label{fig:motivation}
\end{figure}

What we need, instead, is a method that estimates the scene and humans from images of a single color camera. 
This is challenging, as the lack of depth information causes the scale and placement of objects to be inconsistent \wrt the humans interacting with them.
This leads to physically implausible results, like humans penetrating objects, or lacking physical contact when walking, sitting, or lying down, causing bodies to ``hover'' in the air (see \Cref{fig:motivation}).
Methods that reconstruct \threeD humans from single views leverage statistical body models \cite{Joo2018_adam, SMPL:2015, pavlakos2019expressive, xu2020ghum} as priors on the body shape and pose.
However, the same tools do not exist for the collective space of \threeD scene layouts. 
This is due to the enormous space of possible object arrangements in indoor \threeD scenes, the large number of different object classes, and the huge inter-class (\eg, chairs and desks) and intra-class (\eg, desk chair and club chair) shape variability.

To address the above issues, we present \modelname, which stands for \modelnameLong. 
\modelname leverages information across several \HSI frames to estimate both a plausible \threeD scene and a moving human that interacts with the scene. 
Figure~\ref{fig:teaser} provides a high-level overview. 
\modelname takes as input: 
{(1)}     a set of color frames from a static monocular camera, 
{(2)}     a \threeD human mesh inferred for each frame \cite{pavlakos2019expressive,Kocabas_PARE_2021}, and
{(3)}     a \threeD shape inferred for each object detected in the scene~\cite{nie2020total3dunderstanding,kirillov2020pointrend}. 
As output, \modelname produces a refined \threeD scene, comprised of repositioned input objects, so that it is consistent with the estimated \threeD human; \ie, it satisfies the expected contacts on the body \cite{hassan2020populating}, while preventing interpenetration.
\modelname uses a novel  optimization scheme, that jointly optimizes over camera pose, ground-plane pose, and the size and position of \threeD objects, while being constrained by various \HSI constraints.

\modelname takes three types of \HSI constraints into account:
(1) humans who move in a scene are occluded or occlude objects, thus, defining the depth ordering of the objects  (c.f.~\cite{Essa2001}),
(2) humans move in free space that is not occupied by objects and do not interpenetrate objects,
(3) contact between humans and objects means that the contacting parts of their surfaces occupy the same place in space.
Thus, we leverage both explicit (\ie, contact) and implicit (\ie, free space, no penetrations) \HSI cues.
\modelname is able to use these because it employs detailed meshes for both the scene and the moving human.
In contrast, the few attempts that have been made in this direction use oversimplified shapes \cite{chen2019holistic++}, \ie, \threeD bounding boxes for objects and skeletons for humans, work only for static humans that contact a single object \cite{zhang2020perceiving}, or do not integrate information across several interaction frames \cite{weng2020holistic,zhang2020perceiving, chen2019holistic++}.
Comparisons against the state of the art on the \prox~\cite{hassan2019resolving} and \pigraphs~\cite{savva2016pigraphs} datasets show, that \modelname estimates more accurate and realistic \threeD scene layouts that satisfy the expected contacts, while minimizing penetrations, \wrt the moving humans. 
Interestingly, we find that \modelname's estimated \threeD scene can be used to refine the human poses, with a \prox-like method \cite{hassan2019resolving}.
While estimating \threeD scenes and humans from a single camera is challenging, our results suggest that they are synergistic tasks that benefit each other.

\section{Related Work}
\qheading{Single-view \threeD Human Pose in ``Isolation'':} 
Estimating human pose from an image is a long standing problem \cite{Review_Moeslund_2006,Sarafianos:Survey:2016}. 
Typically, this is cast as estimating \twoD or \threeD joints of 
body        \cite{andriluka2018poseTrack,martinez_2017_3dbaseline,rogez2016mocap,bugra_bmvc_2016,tome2017lifting} or
whole-body skeletons \cite{OpenPose_PAMI,jin2020zoomNet,weinzaepfel2020dope}.
Recently, there has been a significant shift in research interest towards reconstructing the \threeD human body surface which, in contrast to the joints, interacts directly with objects and can be observed by commodity cameras.
To this end, many non-parametric methods 
\cite{Gabeur_2019_ICCV,kolotouros2019convolutional,Saito_2019_ICCV,Saito_2020_CVPR,Smith_2019_ICCV,varol2018bodynet,Zheng_2019_ICCV,xiu2022icon}
have been developed, that estimate either
depth maps                  \cite{Gabeur_2019_ICCV,Smith_2019_ICCV},
\threeD voxels              \cite{varol2018bodynet,Zheng_2019_ICCV}, 
\threeD distance fields     \cite{Saito_2019_ICCV,Saito_2020_CVPR}, or
free-form \threeD meshes    \cite{kolotouros2019convolutional}.
While these methods can reconstruct bodies with details like hair and clothing, 
they do not encode body parts or provide correspondence across people and poses.
%
In contrast, parametric statistical  \threeD shape models of the body \cite{Anguelov05,hasler2009statistical,SMPL:2015} or body, face, and hands \cite{Joo2018_adam,pavlakos2019expressive,romero2017embodied,xu2020ghum} provide this information and  allow  re-posing.
Since parametric models represent the shape and pose in a low-dimensional space, they are a powerful tool to estimate the surface from incomplete data (e.g., 2D images with occlusions) through optimization \cite{bogo2016keep,Joo2018_adam,pavlakos2019expressive,Xiang_2019_CVPR}, 
regression \cite{Choutas2020_expose,Kanazawa2018_hmr,kocabas2020vibe,SPIN:ICCV:2019, khirodkar_ochmr_2022}, or hybrid approaches \cite{joo2020eft}.
\begin{figure*}
    \centering
    \includegraphics[width=\textwidth]{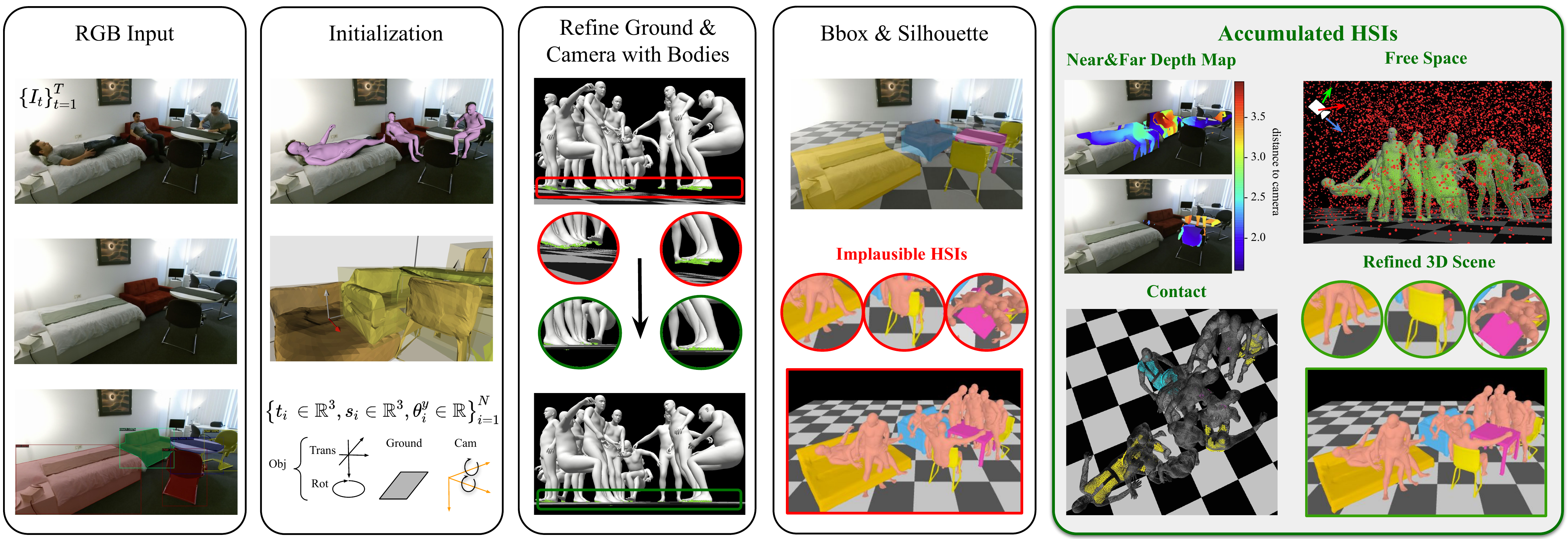}
    \caption{
                Overview of \modelname. 
                Given a video/multiple images, the initialization involves
                using \cite{nie2020total3dunderstanding} to reconstruct a \threeD scene from labeled or detected \twoD instance segmentation masks \cite{kirillov2020pointrend}, 
                estimating the \threeD human poses and shape \cite{pavlakos2019expressive,Kocabas_PARE_2021}, and 
                extracting the expected contact vertices on the estimated bodies using \posa \cite{hassan2020populating}. 
                The first step then refines the camera orientation and ground plane using the human bodies and their foot contact. 
                Then we optimize the object layout based on \twoD bounding boxes and silhouettes to remove interpenetration between people and objects, \eg, the human sits through the chair, stands in a table, or the legs penetrate a bed. 
                Finally, incorporating multiple {\HSI}s collectively from the whole video, we improve the \threeD scene further such that the bodies perform more realistic {\HSI}s.
    }
    \label{fig:pipeline}
\end{figure*}

However, all the above methods reason about the human in ``isolation'', \ie~without taking the surrounding objects and scenes into account. 
Thus, they struggle to reconstruct details like contact with objects, and often fail due to occlusions (\eg, bodies standing behind furniture).
PARE \cite{Kocabas_PARE_2021} addresses this by leveraging localized features and attention, gaining robustness to occlusions.
We initialize our approach with \cite{Kocabas_PARE_2021} to refine the \threeD scene layout.

\smallskip
\noindent
\textbf{Single-view \threeD Scene in Isolation:} 
\threeD reconstruction from single views has been addressed in several recent works that leverage learned geometric priors for specific object classes or entire scenes.
Shapes from single views are reconstructed using generative models for specific object classes \cite{choy20163d,atlasnet,wang2018pixel2mesh,mescheder2019occupancy,srns}.
The methods differ in the underlying representation, which ranges from volumetric representations like occupancy fields~\cite{mescheder2019occupancy} and implicit surface functions~\cite{park2019deepsdf,liu2020dist}, to explicit surface representations like triangular meshes~\cite{wang2018pixel2mesh,gkioxari2019mesh}.
To reconstruct scenes, single objects can be detected~\cite{he2017mask} and reconstructed in isolation.
Mesh-RCNN~\cite{gkioxari2019mesh} detects the objects in an RGB image, and predicts geometry for each object individually.
Instead of a generative mesh model, Izadinia et al.~\cite{izadinia2017im2cad} and Kuo et al.~\cite{kuo2020mask2cad} retrieve individual CAD models for the detected objects in the scene.
Bansal et al.~\cite{bansal2016marr} infer a normal map from the input image that is used to align a retrieved CAD model.
Instead of predicting normal maps from the input image, several methods estimate depth maps~\cite{laina2016deeper,godard2017unsupervised,fu2018deep,shin20193d}, or pixel-aligned implicit functions for objects \cite{Saito_2019_ICCV,Saito_2020_CVPR,xiu2022icon} and scenes \cite{denninger20203d,dahnert2021panoptic}.
Joint estimation of the room layout and objects with scene context information
is done for isolated \threeD scenes without humans in them
\cite{choi2013understanding,huang2018cooperative,huang2018holistic,zhang2017physically,zhao2013scene,Zhang_2021_CVPR,nie2020total3dunderstanding}.
%
%
%

%
Note that there are also methods that predict room layouts with \threeD bounding boxes  \cite{dasgupta2016delay,hedau2009recovering,lee2009geometric,mallya2015learning}.
In contrast, we reconstruct the detailed object geometry to leverage explicit contact point constraints based on the human scene interactions, while optimizing for the scene layout.
%

\begin{table}[t]
    \centering
    \footnotesize
    \resizebox{\linewidth}{!}{
    \begin{tabular}{l|ccccc}
    \toprule
    \tabincell{l}{Method} & \tabincell{c}{GDI} & \tabincell{c}{Cam.} & \tabincell{c}{C-HOI} & \tabincell{c}{N-HOI} & \tabincell{c}{FGC} \\ \hline
    \phosa \cite{zhang2020perceiving} & \textcolor{green}{\cmark} & \textcolor{red}{\xmark} & \textcolor{green}{\cmark}  & \textcolor{red}{\xmark}   & \textcolor{red}{\xmark} \\ \hline
    
    \holisticplus \cite{chen2019holistic++}& \textcolor{red}{\xmark} & \textcolor{red}{\xmark}  & \textcolor{red}{\xmark} & \textcolor{red}{\xmark} & \textcolor{green}{\cmark}  \\ \hline
    \holisticmesh \cite{weng2020holistic} & \textcolor{green}{\cmark} & \textcolor{green}{\cmark} & \textcolor{green}{\cmark} & \textcolor{red}{\xmark} & \textcolor{green}{\cmark} \\ 
    \midrule
    \bottomrule
    \textbf{Ours} & \textcolor{green}{\cmark} & \textcolor{green}{\cmark}  & \textcolor{green}{\cmark} & \textcolor{green}{\cmark} & \textcolor{green}{\cmark} \\ \hline
    \end{tabular}
    }
    \caption{Comparison of the most relevant methods. GDI: Geometric Detailed Interaction. 
    C-HOI: Contact-Human-Object Interaction.
    N-HOI: Exploiting free space constraints with no object contact. 
    FGC: Feet-Ground Contact. Cam.: Camera orientation and ground-plane are refined with humans or not.} 
    \label{differences_in_related_work}
\end{table}

\smallskip
\noindent
\textbf{\threeD Human-Scene Interaction:} 
Humans inhabit \threeD scenes.
Several methods model this and learn to populate a \threeD scene \cite{hassan2020populating,Li2019puttingHumansScenes,PLACE:3DV:2020,zhang2020generating}.
In contrast, our work reasons about the human and its interaction with the \threeD scene from RGB observations.
There are several methods that explore different kinds of 
\hsi;
these can be divided into three categories by the interaction granularity between the human and scene:
(1) Hand-Object~\cite{cao2021reconstructing,yang2021cpf,chao2021dexycb,liu2021semi,jiang2021hand,grab,kwon2021h2o}. 
(2) Body-Object~\cite{zhang2020perceiving,GraviCap2021,taheri2021goal,Black_TrackPople,Laptvev_CVPR_2019_forces}. 
(3) Body-Scene~\cite{iMapper2018,chen2019holistic++,weng2020holistic,Huang:CVPR:2022,hassan2019resolving,savva2016pigraphs}. 
Our proposed method focuses on reconstructing \threeD scenes composed of objects and structural elements like the floor plane, using accumulated human scene interactions (body-objects and body-scene).
Table~\ref{differences_in_related_work}, overviews the most related work that operates on single-view \rgb images/videos.
\phosa \cite{zhang2020perceiving} infers humans and objects together when they are in contact. They do not consider the fact that humans do not need to contact an object to constrain its location; their movement through free space constrains object placement.
Zanfir \etal \cite{zanfir_2018_cvpr} only consider feet-ground contact.
\iMapper \cite{iMapper2018} maps \rgb videos to dynamic ``interaction snapshots'', by learning ``scenelets'' from \pigraphs data and fitting them to videos. However, the estimated scene is not aligned with the \twoD image, and consists of pre-defined CAD templates with fixed shape and size.
\holisticplus \cite{chen2019holistic++} takes learned \threeD \hoi (Human Object Interaction) into account to jointly reason about the arrangement of bodies and objects.
Both \cite{iMapper2018} and \cite{chen2019holistic++} do not model geometrically detailed human-scene interaction, due to their simplified representation of the scene and bodies.
Weng et al.~\cite{weng2020holistic} jointly optimize the reconstructed mesh-based \threeD scene and bodies, which are initialized from \cite{nie2020total3dunderstanding} and \cite{pavlakos2019expressive}.
The approach only considers interpenetration between objects and the human, and does not model the explicit human-scene contact.
Additionally, both~\cite{weng2020holistic,chen2019holistic++} do not model the coherence of human-scene interactions across frames from monocular video.
In contrast to the prior work, our contribution lies in incorporating multiple human-scene interactions collectively, such that we can reconstruct a more accurate and consistent scene, with physically plausible human-scene interactions.

\section{Method}
\label{sec:method}

\modelname is an optimization-based approach that reconstructs a physically plausible 3D scene that is consistent with predicted human-scene interactions over time (see \Cref{fig:pipeline}).
Specifically, our method takes an RGB video or multiple images $\left\{I_{t}\right\}_{t=1}^{T}$ as input and reconstructs the human bodies at each time step $t$ as well as the numerous static scene objects, all of which reside in a common 3D space and are supported by a ground plane.
In our experiments, we consider indoor scenes containing large objects with which humans frequently interact, i.e., chairs, beds, sofas, and tables.
We initialize our approach using separate estimates for the 3D human poses~\cite{Kocabas_PARE_2021,pavlakos2019expressive}, the 3D scene~\cite{nie2020total3dunderstanding}, and the ground plane.
Using the estimated body poses, we predict contact vertices $\mathcal{C}$ for all bodies using \posa~\cite{hassan2020populating}, which predicts likely contact vertices on the body conditioned on pose.
We further divide these vertices into foot contacts $\mathcal{C}^\text{feet}$ and other body part contacts $\mathcal{C}^\text{body}$. 
The explicit foot contact points $\mathcal{C}^\text{feet}$ are used as constraints to refine the camera orientation and ground plane prediction.
Based on this initialization, we optimize the alignment of the objects by minimizing an objective function based on multiple human-scene interactions (HSIs) across the entire input data.
%


\subsection{3D Scene Layout Optimization}
\label{hsi_scene}

Our method leverages multiple {\hsi}s to refine the 3D scene.
Recall that these {\hsi}s provide the following constraints:
(1) humans that move in a scene are occluded or occlude objects, thus, defining the depth ordering of the objects (depth order constraint),
(2) humans move through free space and do not interpenetrate objects (collision constraint),
(3) when humans and objects are in contact, the contact surfaces occupy the same place in space (contact constraint).
Using these constraints, our objective is: 
\begin{multline}\label{e_human_scene}
\mathcal{L}_\text{scene-human}= \lambda_{1}\mathcal{L}_\text{bbox} + \lambda_{2}\mathcal{L}_\text{occ-sil} + \lambda_{3}L_\text{scale} \\
+ \lambda_{4}\mathcal{L}_{\text {depth}}+\lambda_{5}\mathcal{L}_\text{collision}+\lambda_{6}\mathcal{L}_\text{contact}.
\end{multline} 
We apply an occlusion-aware silhouette term $\mathcal{L}_{\text{occ-sil}}$ from \cite{zhang2020perceiving}, a \twoD bounding box projection term $\mathcal{L}_\text{bbox}$ that constrains the top-left corner and the width of the bounding boxes of the objects, and $L_\text{scale}$, 
an $\ell_2$ regularizer to constrain object-scale variation,
\supmat 

\begin{figure}
    \centering
    \includegraphics[width=\columnwidth]{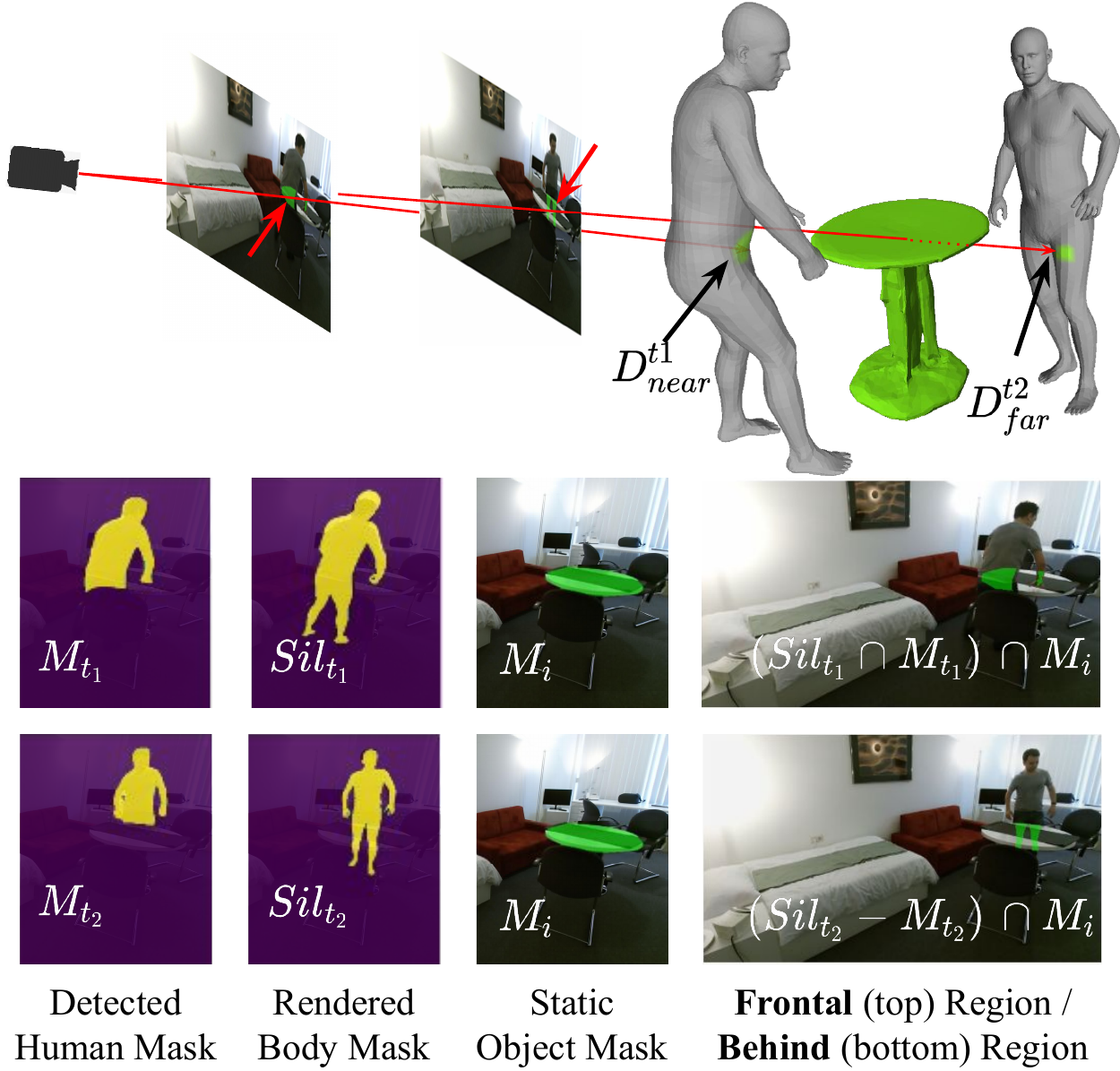}
    \caption{Computing depth range maps for the depth order constraint $\mathcal{L}_\text{depth}$.
    Given a detected human mask $M_{t}$ and a rendered body mask $Sil_{t}$, for each object $i$, we compute the overlap region between $M_{i}$ and $Sil_{t} \cap M_{t}$ as the frontal region and  extract the depth of the backside surface of the body as the near depth range $D_{near}^{t}$ of the object $i$. Similarly, we compute $(Sil_{t}-M_{t}) \cap M_{i}$, which defines the far depth range $D_{far}^{t}$ of the object.
     }
    \label{fig:depth_illustration}
\end{figure}

\paragraph{Depth Order Constraint $\mathcal{L}_\text{depth}$.} 
The occlusion between humans and objects can provide clues about the object's depth.
We assume the human's depth is accurate.
If a human occludes an object, then the far side of the person sets a limit on how close the object can be.
Alternatively, if the object occludes the person, then the visible side of the person sets a maximum distance for the object.
This is summarized in \fref{fig:depth_illustration}.  
In this way, human-object occlusion provides constraints on scene layout even when there is no human-object contact.

Directly applying the ordinal depth loss proposed by Jiang \etal~\cite{jiang2020coherent} for each image
is inefficient, as the required memory increases with the number of images.
In contrast, we accumulate all single depth ordering maps into one far depth range map $\hat{D}_\text{far}$ and one near depth range map $\hat{D}_\text{near}$ as:
\begin{equation*}
    \begin{split}
    \hat{D}_\text{far}(p)&=\min\left (D_\text{far}^{1} (p),...,D_\text{far}^{T} (p)\right), \\
    \hat{D}_\text{near}(p)&=\max\left (D_\text{near}^{1} (p),...,D_\text{near}^{T} (p)\right),
    \end{split}
\end{equation*}
where the pixel $p$ is in the overlapping region between the human bodies and the objects.
Using these accumulated depth range maps, we constrain the depth $D_i(q)$ of a projected pixel $q$ from object $i$ to lie in the corresponding range:
\begin{equation*}
    \begin{split}
    \mathcal{L}_\text{depth} = \sum_{i} \sum_{q \in Sil_{i} \cap M_{i}}  \lbrack &\text{ReLU}(D_{i}(q)-\hat{D}_\text{far}(q)) \\ +  &\text{ReLU}(\hat{D}_\text{near}(q) -D_{i}(q)) \rbrack,
    \end{split}
\end{equation*}
where $Sil_{i}$ is the rendered silhouette of the object $i$, $M_{i}$ is its \twoD segmentation mask, and $D_{i}(q)$ is the depth of the object $i$ at the pixel $q$. \Supmat  

\paragraph{Collision Constraint $\mathcal{L}_\text{collision}$.} 
To penalize all interpenetrating vertices of objects and bodies in the scene, we use the signed distance field (SDF) of all reconstructed bodies.
Specifically, we calculate a signed distance field volume $V_{j}$ for each body $j$ in a shared 3D world space, and accumulate them into a global SDF volume as $\hat{V} = \min (V_1,...,V_j,...)$.
The SDF $\hat{V}$ is stored in a volumetric grid of size $256^3$, which spans a padded bounding box of all bodies.
For a vertex $u_{i}$ of an object $O_{i}$, we compute the voxel coordinates $f(u_{i})=(p(u_{i}), q(u_{i}), k(u_{i}))$ in the global SDF volume, 
and retrieve the corresponding SDF value $\hat{V}_{f(u_{i})}$.

Based on the SDF values of all vertices of all $N$ objects, we resolve the scene-body interpenetration by penalizing vertices with a negative SDF value:
\begin{equation*}
    \vspace{-5pt}
    \mathcal{L}_\text{collision}=\sum_i \sum_u\lVert \hat{V}_{f(u_{i})}\rVert^{2}_{2}, 
    ~~~
    \hat{V}_{f(u_{i})}<0.
    \vspace{-4pt}
\end{equation*}

\paragraph{Contact Constraint $\mathcal{L}_{\text{contact}}$.} 
When humans and objects are in contact, the contact surfaces occupy the same place in space.
We propose a contact constraint to minimize the distance between the contacted body parts and its assigned corresponding contacted object.
\phosa \cite{zhang2020perceiving} proposes a loss in which they assign a whole body to only one object, whereas humans sometimes  interact with multiple objects; e.g., a person sits on a chair and puts their hand on a table.
In contrast, we directly assign the contacted body vertices $\mathcal{C}^{body}_{i}$ of each body to different objects, based on the overlap between the \twoD projection of the vertices and the detected object masks, and based on the \threeD distances between them.
We consider the vertices of sofa and chair backs and seat bottoms as contactable regions, \supmat

We minimize the distance between the contacted bodies and the contacted object parts: 
%
\begin{align*}
    \hspace{-1.0 em}
    \mathcal{L}_{\text{contact}}=
    \sum_{i} \sum_{v \in \mathcal{C}^\text{body}}  \mathbb{I}(v, O_{i})
    \lbrack &\text{CD}(v^{y},\mathcal{C}(O_i)^{y}) \\
    + &\text{CD}(v^{\perp y},\mathcal{C}(O_i)^{\perp y})  \rbrack ,
\end{align*}
where $\mathcal{C}(O_i)^{\perp y}$ and $\mathcal{C}(O_i)^{y}$ denote the back and the bottom seat contact part of an object $i$, respectively, 
$y$ denotes the y-axis direction and $\perp y$ the vertical direction to it.
$\mathbb{I}(v, O_{i})$ is an indicator function ($1$ only if the contact vertex $v$ is assigned to the contacted object $O_i$, 0 else).
$\text{CD}$ denotes the one-directional Chamfer Distance (CD), i.e., from bodies to objects, because for large furniture like a bed or a sofa, a human only contacts a small region of the object.
In contrast, \phosa \cite{zhang2020perceiving} uses a bi-directional CD, which tends to shrink the object to match the contacted body parts.
%


\subsection{Optimization}
We optimize \eref{e_human_scene} for a specific scene w.r.t. the parameters $\mathbf{s}_{i}$ (scale), $\mathcal{\theta}_{i}$ (rotation), $\mathbf{t}_{i}$ (translation) of the objects $\{{i=1...N}\}$, with the Adam optimizer~\cite{kingma2014adam}.
In the following, we detail the initialization of the 3D scene and the HPS.

\paragraph{Initial 3D Scene.}
We extract a representative 2D image $\mathbf{I}$ from the input data without any human-object occlusion.
For this image, depending on the experiment, we either use the ground truth  
2D bounding boxes ${B}_{i}$ and instance masks ${M}_{i}$ for all $N$ objects in the scene or compute them using PointRend~\cite{kirillov2020pointrend}.
We use \cite{nie2020total3dunderstanding} to get an initial 3D scene $\mathbf{S}_{0}$, consisting of a ground plane $y=y_{gp}$ and multiple object meshes $\{O_{i}\}_{i=1}^{N}$, and a perspective camera with \textsl{roll} and \textsl{pitch}.
Each object $i$ has a translation $\mathbf{t}_{i} \in \mathbb{R}^{3}$, scale $\mathbf{s}_{i} \in \mathbb{R}^{3}$, and a rotation around 
the y-axis $\theta_{i}^y \in [0, 2\pi)$.
Since the predicted meshes of~\cite{nie2020total3dunderstanding} are incomplete and have holes, we use Occupancy Networks~\cite{mescheder2019occupancy} and Marching Cubes \cite{Lorensen1987MarchingCA} to transform each object mesh into a water-tight mesh.
Based on this preparation, we first optimize the objective function without considering the {\hsi}s:
\begin{equation*}
    \mathcal{L}_\text{scene}=\mathcal{L}_{\text{occ-sil}} + \lambda_{1}\mathcal{L}_\text{bbox}+\lambda_{2}\mathcal{L}_\text{scale}.
\end{equation*}
%

\paragraph{Initialization of the ground and camera.}
As shown in the third column of \Cref{fig:pipeline}, the estimated ground plane and camera orientation from \cite{nie2020total3dunderstanding} are inconsistent with the reconstructed bodies (\eg, people float in the air). 
Previous methods either fix the camera orientation and only optimize the ground plane and humans \cite{chen2019holistic++}, or estimate them independently per image \cite{weng2020holistic}, which generates inconsistent camera orientations and ground planes throughout a video.
However, the camera orientation and ground plane are essential for producing plausible {\hsi}s.
Thus, we jointly estimate the ground, camera and multiple humans together, by applying:  
\begin{equation*}
    \mathcal{L}_\text{feet}(R,p) = \rho({R}^{\top}\sum_{t}{\mathcal{C}_{t}}^\text{feet} - [0,y_{gp},0]^\top; \sigma_1),
\end{equation*}
where ${R}$ is the camera rotation matrix calculated from \textsl{pitch}, and \textsl{roll}, and $\rho$ denotes a robust Geman-McClure error function~\cite{robustifier} for down-weighting outliers and $\sigma_1=0.1$.

\paragraph{Initial Estimate of 3D Bodies.}
To obtain an initial body shape and pose estimate for the input images $\left\{I_{t}\right\}_{t=1}^{T}$, 
we use \openpose \cite{OpenPose_PAMI} and \smplifyx \cite{pavlakos2019expressive}.
Specifically, we use a perspective camera  and estimate the pose parameters $\theta_{t}$ of \smplx for each frame with shared body shape parameters $\beta$. 
\smplifyx requires a good initialization and, for this, we use \pare \cite{Kocabas_PARE_2021} because it is robust to occlusion and our scenes involve significant occlusion.
\pare outputs \smpl, which we convert to \smplx~\cite{smpl2smplx}, and use the resulting \threeD joints to initialize {\smplifyx}, \supmat

We then optimize all {\smplx} parameters to minimize an objective function $E_\text{Body}$ of multiple terms, as described in \smplifyx \cite{pavlakos2019expressive} (see {$E_{\text{SMPLify-X}}$}) :
\begin{equation*}
    E_\text{Body} = \sum_{t=1}^{T}\left(E_{\text{SMPLify-X}}\left(t\right)\right) + \lambda_\text{smooth}\mathcal{L}_\text{smooth}.
\end{equation*}
To reduce jitter, we add a constant-velocity motion smoothing term on 3D joints $J$ and their 2D projections $J^\text{Proj}$:
\begin{align*}
    \mathcal{L}_\text{smooth}=\sum_{t=1}^{T-1}
    &\rho\left(\lVert J_{t-1}+J_{t+1}-2 \times J_{t} \rVert; \sigma_2\right)\\ +  
    &\rho\left(\lVert J_{t-1}^\text{Proj} + J_{t+1}^\text{Proj} - 2 \times J_{t}^\text{Proj} \rVert; \sigma_3 \right),
\end{align*}
where $\sigma_2=0.1$ and $\sigma_3=100$.
To avoid noisy and unreliable body poses, and therefore, incorrect human-scene interactions during optimization, we also 
filter out outliers
based on a constant-velocity assumption.
To that end, we calculate the acceleration of the pelvis $\nu_{t}$ and the joints $\alpha_{t}$ of a person in frame $t$. 
%
We filter out frames in which either pelvis translation or joint velocities are above a threshold; that is, 
$
     \{j: \nu_{j}<\tau_\text{pelvis} \cap \alpha_{j} < \tau_\text{local} \mathrm{, }\,\, j \in  \{1...T\}\}
$,
where $\tau_{pel},\,\, \tau_{local}$ are the thresholds for the pelvis acceleration and the local pose acceleration, respectively.

\section{Experiments}
To evaluate the influence of accumulated {\hsi}s on the optimized 3D scene layout, we use two different datasets, \pigraphs~\cite{savva2016pigraphs} and \prox~\cite{hassan2019resolving} (see Sup.~Mat.).
In comparison to \cite{nie2020total3dunderstanding} and \cite{weng2020holistic}, we achieve \stateoftheart 3D scene layout reconstruction, both quantitatively (see \Cref{sec:results_quan}) and qualitatively (see \Cref{sec:results_qual}).
On the \prox \emph{quantitative} dataset, we find that our 3D scene reconstructions lead to more accurate human shape and pose estimations than our baselines.
In \Cref{sec:results_ablation}, we analyze the different energy terms and how they contribute to our final results.
%
%

\begin{table*}[t]
\centering
\footnotesize
\resizebox{\linewidth}{!}{
\begin{tabular}{c|c|c|ccc|ccc|cc}
    \toprule
     Methods & \multicolumn{5}{c}{Setting}{} & \multicolumn{3}{c}{Scene Recon.} & \multicolumn{2}{c}{HSI} \\ 
     \hline
      & BBOX\&Mask & Cam. & Contact & Depth & Colli.  & $\text{IoU}_\text{3D}$ $\uparrow$ & P2S$\downarrow$ &  $\text{IoU}_\text{2D}$ $\uparrow$ & Non-Col $\uparrow$ & Cont. $\uparrow$ \\ \hline 
     HolisticMesh~\cite{weng2020holistic} & PointRend  &&&& & 0.211 & 0.410 &0.648  & 0.990 & 0.369  \\
     Total3D~\cite{nie2020total3dunderstanding} & PointRend &&&&& 0.246  &0.319 &0.522& 0.974& 0.510 \\   
     \rowcolor{lightgray}
     \textbf{Ours} & PointRend & \textcolor{GreenColor}{\cmark} & \textcolor{GreenColor}{\cmark} & \textcolor{GreenColor}{\cmark} & \textcolor{GreenColor}{\cmark}  & 0.309 & 0.221 & 0.777  & 0.977 & 0.612  \\ 
         \midrule
     HolisticMesh~\cite{weng2020holistic} &  \twoD GT  &&&&& 0.267 & 0.237 & 0.745  & \textcolor{red}{0.988} & 0.491 \\
     Total3D~\cite{nie2020total3dunderstanding} &  \twoD GT &&&&         & 0.196 & 0.369 &  0.227 & 0.963 & 0.440 \\   
     \rowcolor{lightgray}
    \textbf{Ours} & \twoD GT & \textcolor{GreenColor}{\cmark} & \textcolor{GreenColor}{\cmark} & \textcolor{GreenColor}{\cmark} & \textcolor{GreenColor}{\cmark}  &0.383 & \textcolor{blue}{0.199} & {0.898} & \textcolor{blue}{0.986} & 0.673 \\ 
    
    \midrule
    &  & \textcolor{GreenColor}{\cmark} & & & & 0.374 & 0.206 & 0.859 & 0.979 & \textcolor{blue}{0.738}  \\ 
    
    &  & \textcolor{GreenColor}{\cmark} & \textcolor{GreenColor}{\cmark} & \textcolor{GreenColor}{\cmark} &   & \textcolor{blue}{0.389} & \textcolor{blue}{0.199} & \textcolor{blue}{0.904} &	0.983 &	0.697 \\
    
    Ablation Study &  \twoD  GT & \textcolor{GreenColor}{\cmark} & \textcolor{GreenColor}{\cmark} & & & 0.381 & 0.205 & \textcolor{blue}{0.904} & 0.980 & \textcolor{red}{0.773}  \\
    
     &   & \textcolor{GreenColor}{\cmark} & & \textcolor{GreenColor}{\cmark}   && \textcolor{red}{0.393} & \textcolor{red}{0.194} & \textcolor{red}{0.907} &	0.983 &	0.638 \\
     
      &  & \textcolor{GreenColor}{\cmark} &  & &  \textcolor{GreenColor}{\cmark}    & {0.383} & \textcolor{blue}{0.199} & 0.903 &	0.984 &	0.674  \\

    \bottomrule
    
\end{tabular}
}
\caption{Quantitative results for \threeD scene understanding (\threeD object detection) and human-scene interaction on the  \prox \textsl{qualitative} dataset. P2S, Non-Col and Cont denote \textsl{point2surface distance}, Non-Collision and Contactness respectively. In each column, \textcolor{red}{red} is the best result among methods that take \twoD labeled masks as input; \textcolor{blue}{blue} is the second best.
The check marks indicate which constraints are used.}
\label{tab:quan_scene}
\end{table*}

\begin{table}[t]
    \centering
    \footnotesize
    \begin{tabular}{lcc}
    
    \toprule

    Methods & $\text{IoU}_\text{2D}$ $\uparrow$ & $\text{IoU}_\text{3D}$ $\uparrow$ \\ \hline

    Cooperative \cite{huang2018cooperative} & 68.6 & 21.4 \\
    Holistic++ \cite{chen2019holistic++} & 75.1 & 24.9 \\
    HolisticMesh \cite{weng2020holistic} & 75.6 & 26.3  \\ \hline
    \textbf{Ours} & \textbf{79.2} & \textbf{27.8} \\
    \bottomrule
    
    \end{tabular}
    \caption{Quantitative results for \threeD scene understanding (\threeD object detection) on \textsl{\pigraphs} dataset~\cite{savva2016pigraphs}.}
    
    \label{tab:quan_scene_pigraph}
\end{table}

\begin{table}[t]
    \centering
    \footnotesize
    \resizebox{\linewidth}{!}{
    \begin{tabular}{lccccc}
    
    \toprule
     & \multicolumn{3}{c}{Cam. Orien.} & \multicolumn{2}{c}{Ground Pen}\\ \hline 
    Methods & $\text{pitch}$ $\downarrow$ & $\text{roll}$ $\downarrow$ & mean $\downarrow$ & Freq.~$\downarrow$ & Dist.~$\downarrow$ \\ \hline
    
    Total3D \cite{nie2020total3dunderstanding} & 0.059 &\textbf{0.031} & 0.045 & 0.316 & 0.167 \\
    \hline
    \textbf{Ours} & \textbf{0.042} & 0.034 & \textbf{0.038} & \textbf{0.100} & \textbf{0.112} \\
    \bottomrule
    \end{tabular}
    }
    \caption{Errors in the camera orientation and the ground penetration using foot contact on the \prox \textsl{qualitative} dataset.}
    \label{quan_camera_ground}
\end{table}


\begin{figure*}[h]
  \centering
    \begin{tabular}{cccc}
        \multicolumn{4}{c}{\hspace{-0.025\columnwidth}\includegraphics[width=0.975\textwidth]{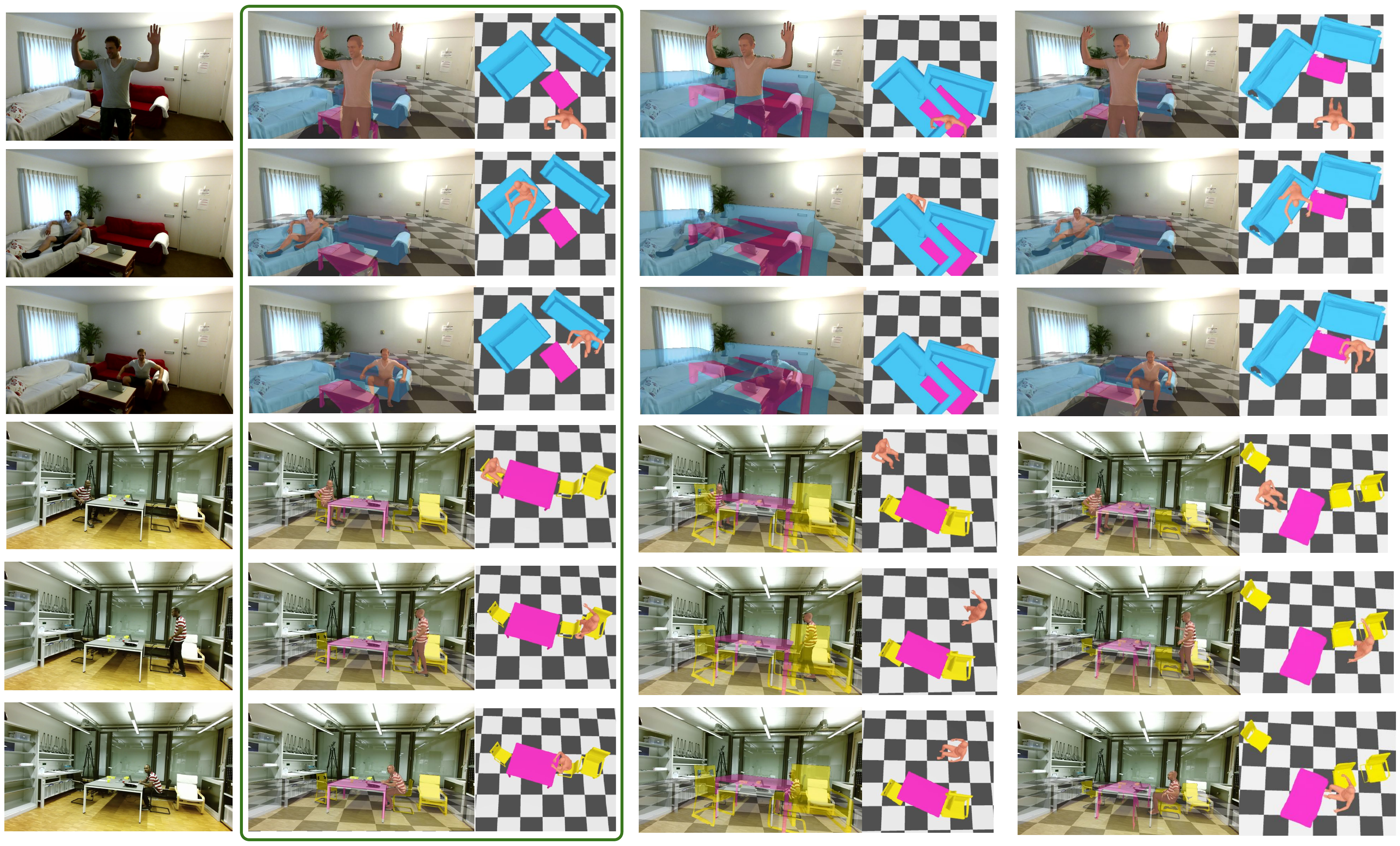}}  
        \vspace{-0.1cm}\\
        \hspace{0.07\columnwidth}  RGB Input & \hspace{0.23\columnwidth} \textbf{Ours} & \hspace{0.37\columnwidth} Total3D \cite{nie2020total3dunderstanding} & \hspace{0.13\columnwidth} HolisticMesh \cite{weng2020holistic}\\
    
    \end{tabular}
  \caption{Qualitative results on \pigraphs~\cite{savva2016pigraphs} (top) and \prox~\cite{hassan2019resolving}~(bottom). 
Our method recovers better \threeD scenes and HPS, which supports more plausible {\hsi}s, compared with two baselines: Total3D \cite{nie2020total3dunderstanding} 
and HolisticMesh \cite{weng2020holistic}. 
  }
  \label{fig:qualitative}
\end{figure*}

\subsection{Quantitative Analysis}
\label{sec:results_quan}

We perform several experiments to investigate the effectiveness of our proposed method in three parts:
\threeD scene reconstruction,
human-scene interaction (HSI) reconstruction, and 
human pose and shape (HPS) estimation.
%
%

\paragraph{\threeD Scene Reconstruction.}
Following~\cite{nie2020total3dunderstanding, huang2018holistic, chen2019holistic++, weng2020holistic}, we compute the \threeD IoU and 2D IoU of object bounding boxes to evaluate the \threeD scene reconstruction and the consistency between the \threeD world and 2D image on \prox and \pigraphs.
However, the \threeD IoU is coarse and does not capture the error in an object's orientation, which is quite important for physically plausible \hsi, e.g., a human can not sit on an armed chair with the wrong orientation.
Therefore, we introduce the \textsl{point2surface distance} ($p2s$) to measure the distance from a cropped object mesh to the estimated 3D object mesh.
It enables 3D scene reconstruction evaluation with more geometric details including orientation and shape.
Given 2D labeled or detected \cite{kirillov2020pointrend} bounding boxes and masks,
our method improves the input \cite{nie2020total3dunderstanding} significantly, and outperforms \cite{weng2020holistic} on all scene-reconstruction metrics and different datasets, as shown in \Cref{tab:quan_scene} and \Cref{tab:quan_scene_pigraph}. 
Furthermore, we evaluate the error of the camera orientation and ground plane penetration \cite{rempe2021humor} using the estimated foot contact vertices (see \Cref{quan_camera_ground}).
We find that  jointly optimizing the camera orientation and the ground plane using foot contact significantly improves accuracy compared to the initial estimate from \cite{nie2020total3dunderstanding}.

\paragraph{Human-scene Interaction Reconstruction.}
To evaluate the 
physical plausibility of the estimated scene,
we compute the metrics used in prior work \cite{zhang2020generating, PLACE:3DV:2020, hassan2020populating}.
Specifically, for each reconstructed body and 3D scene, we calculate
(1) the \emph{non-collision score} to measure the ratio of body mesh vertices that do not penetrate the estimated 3D scene, divided by the number of all body mesh vertices, and 
(2) the \emph{contact score} to denote whether the body is in contact with the 3D scene or not.
The \emph{contact score} is $1$, if at least one vertex of a body interpenetrates  the 3D scene.
We report the mean \emph{non-collision score} and mean \emph{contact scores} among all videos and all bodies.
In \Cref{tab:quan_scene}, \modelname achieves the best balance between non-collision and contact.

The estimated scenes with detected \twoD boxes and masks \cite{kirillov2020pointrend} provide lower \hsi scores than with \twoD GT.
This is mainly because of the mis-detected objects from \cite{kirillov2020pointrend}. 
Since the reconstructed scenes of \cite{weng2020holistic} do not support human-scene contact well, \eg, a sitting body often floats, due to the lack of explicit human-scene contact modeling, it has a better non-collision score but a lower contact score.

\paragraph{Human Pose and Shape (HPS) Estimation.}
Can we use the estimated \threeD scene to, in turn, improve \threeD HPS?
Here we follow \prox but replace the scanned \threeD scene of \prox with our estimated \threeD scene.
In \Tref{quan_body_prox}, we evaluate the HPS estimation on \prox \emph{quantitative} using the metrics from \cite{hassan2019resolving}.
Specifically, we report
(1) the mean per-joint error (PJE) and
(2) the mean vertex-to-vertex distance (V2V).
Unlike the common measures in the field, neither of these metrics align the body with ground truth, either at the pelvis or using full Procrustes alignment.
For completeness, we also compute these metrics with Procrustes alignment, denoted as p.PJE and p.V2V, respectively.
Note that the metrics w./o.~Procrustes alignment (PJE and V2V) are more meaningful here, since we want to evaluate how well the method solves for the translation, rotation, and scaling of the human body.
As shown in \Tref{quan_body_prox}, with estimated camera orientation and ground plane constraints (+CamGP), the PJE and V2V are both improved by a significant margin $\mathrm{+}43.21$ and $+42.41$ respectively, w.r.t. our baseline.
We also see that our refined scene can further refine our estimated bodies by applying the SDF loss (+SDF) and the contact loss (+Contact) from \cite{hassan2019resolving}.
Our final body estimation outperforms \holisticmesh \cite{weng2020holistic} and is similar to \prox, {\em without having access to a scanned 3D scene.}
%


\subsection{Ablation Study}
\label{sec:results_ablation}

To analyze the contribution of the accumulated {\hsi}s and the different constraints,
we conducted multiple ablation studies; see \Cref{tab:quan_scene}.
All three proposed \hsi constraints (depth order, collision, and contact) help to improve 3D scene reconstruction in different ways.
The \emph{contact} constraint produces the highest human-scene contact scores, but decreases the non-collision score. The $\emph{collision}$ and $\emph{depth order}$ both contribute to the non-collision score.
However, using only the $\emph{depth order}$ constraint achieves a slightly better \threeD scene than our full model, but leads to worse human-scene contact scores.
%
By applying all constraints, our method can generate a 3D scene that supports more physically plausible {\hsi}s. 
%


\begin{table}[t]
    \centering
    \footnotesize
    \begin{tabular}{lcccc}
    
    \toprule
    \multicolumn{5}{c}{With \textbf{G.T Captured \threeD Scene Scans}} \\ \hline
    Methods & PJE$\downarrow$ & V2V $\downarrow$& p.PJE $\downarrow$ & p.V2V$\downarrow$\\
    \hline
    RGB~\cite{hassan2019resolving} & 220.27 & 218.06 & 73.24 & 60.80 \\
    \prox~\cite{hassan2019resolving} & 167.08 & 166.51 & 71.97 & 61.14 \\ 
    \midrule
    
    \multicolumn{5}{c}{With \textbf{Image2Mesh Models}} \\ \hline
    HolisticMesh~\cite{weng2020holistic} & 190.78 & 192.21 & \textbf{72.72} & \textbf{61.01}\\
    $\text{baseline}^{*}$ & 219.62& 222.50 &  75.92 & 68.34\\
    
    \text{+CamGP} & 176.41	 & 180.09  & 73.41 &67.33\\

    \text{+CamGP+SDF} & 175.98 & 179.98 &	73.96 & 68.29  \\
    
    \textbf{Ours} &  \textbf{174.37}	& \textbf{178.31} & 73.60 & 67.89 \\

    \bottomrule
    \end{tabular}
    \caption{Quantitative results for human pose estimation on \prox \textsl{quantitative} dataset ($\text{baseline}^{*}$ denotes batch-wise \smplifyx, \textbf{Ours}: +CamGP+SDF+Contact.)}
    \label{quan_body_prox}
\end{table}

\subsection{Qualitative Analysis}
\label{sec:results_qual}
In \fref{fig:qualitative}, we show reconstructed \threeD scenes and humans along with frames from the \rgb videos, to demonstrate the effectiveness and generality of our approach on different datasets (\prox~\cite{hassan2019resolving} and \pigraphs~\cite{savva2016pigraphs}).
%
\modelname recovers better \threeD scenes and HPS compared to Total3D \cite{nie2020total3dunderstanding} 
and HolisticMesh \cite{weng2020holistic}.
See Sup.~Mat.~for more examples.

\section{Discussion}
Based on single-view inputs, our proposed method optimizes the \threeD pose of objects in a scene.
While we assume a static camera, future work should explore moving cameras and structure-from-motion techniques to better estimate the 3D scene.
We also assume that the scene is static. 
However, humans move objects
when interacting with the world, resulting in a dynamic scene layout.
%
%
We believe that our proposed constraints based on {\hsi}s will be beneficial for future work on 
reconstructing
dynamic scenes.
Besides optimizing the \threeD scene layout, we do not change the initial shape estimate of an object.
A more flexible and adjustable geometric object representation, e.g., an implicit representation, would be beneficial.
One could then optimize over the space of object shapes in addition to object poses.
%
While here we focus on large objects like furniture, hand-held objects are also important and are likely subject to different constraints.
During HSI, bodies are often occluded, causing errors in estimated 3D human pose.
These estimates could be improved by incorporating strong human motion priors \cite{Zhang:ICCV:2021, rempe2021humor}.
%
%
%
%

\section{Conclusion}
We have introduced \modelname, which reconstructs a \threeD scene by exploiting \threeD humans interacting with it.
We have demonstrated that accumulated {\hsi}s, computed from a monocular video, can be leveraged to improve the 3D reconstruction of a scene.
The reconstructed scene, in turn, can be used to improve \threeD human pose estimation.
In contrast to the state of the art, \modelname can reconstruct a consistent, physically plausible {\threeD} scene layout.

\medskip

\noindent
{\qheading{Acknowledgments.}
We thank Yixin Chen, Yuliang Xiu for 
the insightful 
discussions, Yao Feng, Partha Ghosh and Maria Paola Forte
for proof-reading, and 	
Benjamin Pellkofer for IT support.
This work was supported by the German Federal Ministry of Education and Research (BMBF): Tübingen AI Center, FKZ: 01IS18039B.
}


{\qheading{Disclosure.}
{\small
\href{https://files.is.tue.mpg.de/black/CoI_CVPR_2022.txt}{
      https://files.is.tue.mpg.de/black/{CoI\_CVPR\_2022.txt}}}


{\small
\bibliographystyle{configs/ieee_fullname}
\bibliography{configs/main}
}
\balance

\newpage
\begin{appendices} \label{appendices}

In this supplemental document, we provide additional information about the dataset, implementation details, extended sensitivity analysis, failure cases, additional qualitative results and discussion of potential misuse.

\section{Dataset} \label{sec:dataset}

\paragraph{\pigraphs.}
\pigraphs \cite{savva2016pigraphs} consists of $60$ RGB-D videos of $30$ scenes.
The dataset is recorded with a \textit{Microsoft Kinect One}, and is designed to capture human and object arrangements in different kinds of interaction.
Each video recording is about $2$-minute long with $5$ fps.
It contains labeled 3D bounding boxes of objects in the scene and human poses represented as 3D skeletons.
We use this dataset to evaluate the scene reconstruction and compare with \cite{nie2020total3dunderstanding, weng2020holistic}.
Note that the provided human poses are noisy and not suitable for an evaluation of 3D human shape and pose estimation.

\begin{figure*}
    \centering
    {\includegraphics[width=\textwidth]{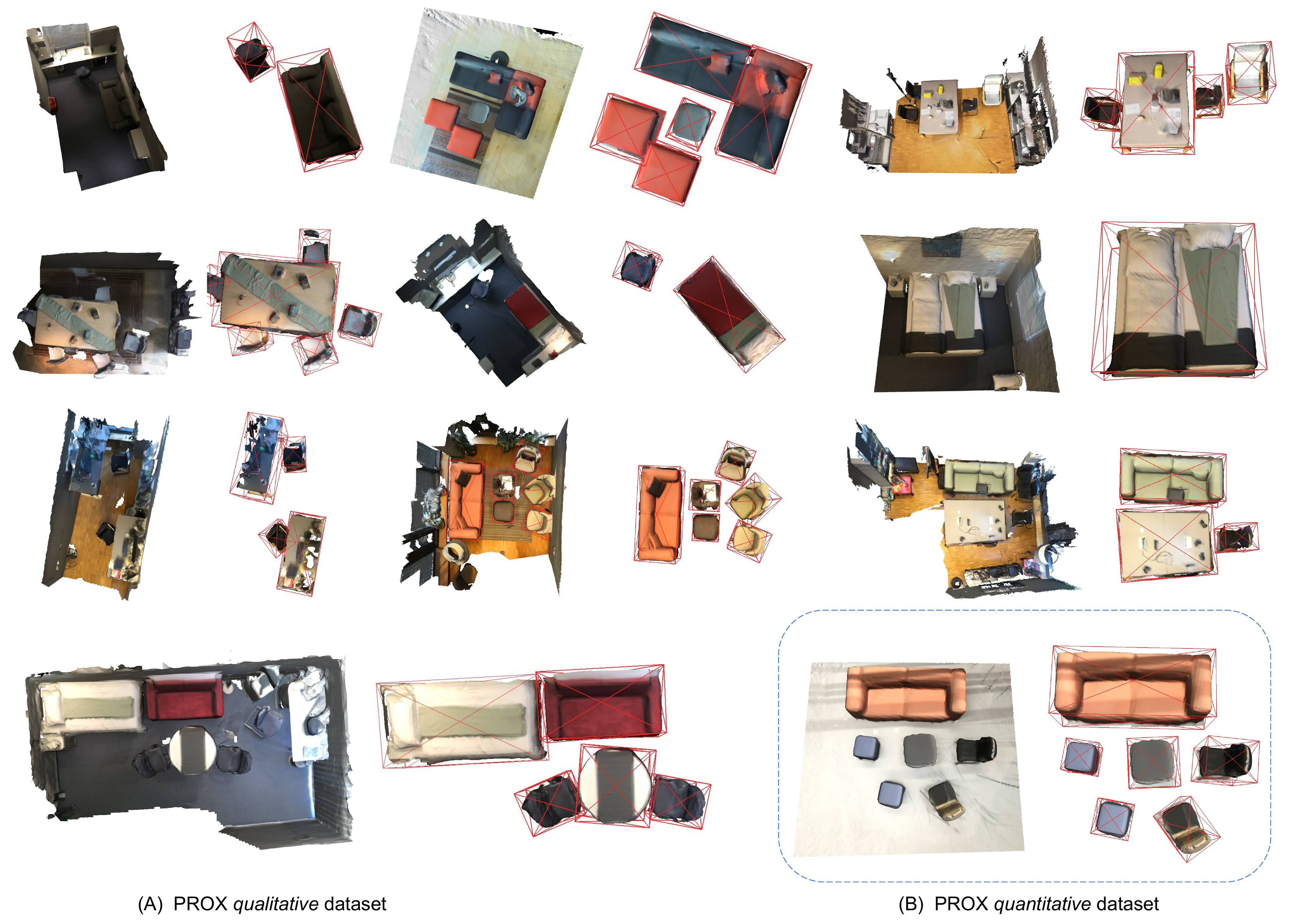}}
    \caption{                   We crop out each object separately and label the corresponding 3D bounding box for 10 scenes in PROX \textsl{qualitative} dataset and one scene in PROX \textsl{quantitative} dataset.
}
    \label{fig:prox_qualitative_dataset}
\end{figure*}

\paragraph{\prox Qualitative.}
\prox \emph{qualitative} contains $61$ RGB-D videos at $30$ fps of human motion/interaction in $12$ scanned static 3D scenes.
The data has been recorded using the \textit{Microsoft Kinect One} and \textit{StructureIO} sensor.
To enable 3D scene reconstruction evaluation on this dataset, we segment and label each object with its \threeD bounding box.
Since there are two scenes (i.e., ``BasementSittingBooth'' and ``N0SittingBooth'') containing an inseparable object, we evaluate all methods on the remaining $10$ scenes (see \fref{fig:prox_qualitative_dataset}) using the corresponding $51$ videos as input. 

\paragraph{\prox Quantitative.}
\prox \emph{quantitative} captures a sequence of human-scene interaction RGB-D frames within a synchronized \textit{Vicon} marker-based motion capturing system.
In total, the dataset contains $178$ frames and provides groundtruth body meshes, which accounts for human pose and shape (HPS) evaluation.
For fair evaluation on HPS, we input all images into \holisticmesh \cite{weng2020holistic} and ours to get a refined scene and use a refined scene to get refined bodies.
In addition, we also label this scene for 3D scene reconstruction evaluation, see \fref{fig:prox_qualitative_dataset}. 

\section{Implementation Details}



\paragraph{Loss Terms.}
The \textit{2D bounding box term} $\mathcal{L}_\text{bbox}$ is 
an $\ell_1$ norm between an object's projected 3D bounding box $\mathit{Proj}_{i}$ and its detected 2D bounding box $\mathit{Det}_{i}$, 
expressed with 
the top-left corner coordinate ${x_{min}, y_{min}}$ and ${width}$ 
value.
\begin{equation*}
\vspace{-5pt}
\mathcal{L}_\text{bbox}=\sum_{i}\|\textsl{Proj}_{i}^{\alpha}-\textsl{Det}_{i}^{\alpha}\|, ~~~\alpha \in \{x_{min}, y_{min}, width\}.
\vspace{-4pt}
\end{equation*}

The \textit{scale term} prevents object scales $s$ deviating far from the initial estimates $s^{init}$ from Total3D~\cite{nie2020total3dunderstanding}:
\begin{equation*}
\vspace{-5pt}
\mathcal{L}_\text{scale} = \sum_{i} \| \frac{s_{i}}{s_{i}^{init}} - 1.0 \|_{2}.
\vspace{-4pt}
\end{equation*}

\paragraph{Initial Estimate of 3D Bodies.}
We use \pare~\cite{Kocabas_PARE_2021} to initialize the body poses and shape (shape $\beta$, pose $\theta$, scale $s$).
Since our approach uses the SMPL-X \cite{pavlakos2019expressive} model, we apply \cite{smpl2smplx} to convert the SMPL parameter estimated from \pare.
In addition, we use perspective projection with the calibrated camera intrinsic parameters, $K$ provided by the datasets (PiGraph and PROX).
To convert the estimations of \pare using a weak perspective camera model, we compute the corresponding translation $t^{body}$ by:
\begin{equation*}
    \Pi_{K_{0}}\left(s(R_{\theta}(J(\beta))\right) = \Pi_{K}\left((R_{\theta}(J(\beta))+t^{body}\right),
\end{equation*}
where $K_{0}$ denotes the camera intrinsic parameters of the weak perspective camera model with focal length 5000.
Then we extract the resulting 3D joints
to initialize $E_{body}$.

\begin{figure}
    \centering
    {\includegraphics[width=\columnwidth]{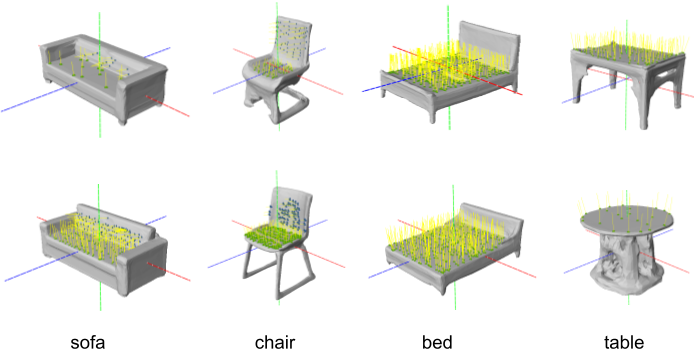}}
    \caption{Contact regions of different objects.
}
    \label{fig:contact_region}
\end{figure}

\paragraph{Contact Regions of Objects.}
We automatically calculate the contact regions of objects based on the normal of the vertices. Specifically, the vertices, whose normals are along y-axis, are the bottom or top part of the objects, while the vertices with along z-axis normal are the back part of the objects. We term that sofas and chairs have two contact regions, i.e., bottom and back parts, while beds and tables only have the top part as the contact region, shown in \fref{fig:contact_region}. 
%

\paragraph{Optimization.}
We use the Adam optimizer \cite{Joo2018_adam} to optimize the final energy term with a step size of $0.002$ and $3000$ iterations.
We set $\lambda_{1}, \lambda_{2}, \lambda_{3}$ as $1000, 0.3, 1000$ respectively, for \twoD bounding box term, occlusion-aware term and scale term.
The weights of our proposed depth order constraint, collision constraint, and contact constraint are set to $\lambda_{4}=8, \lambda_{5}=1000$, and $\lambda_{6}=1e5$, respectively.
%

Our method takes around $30$ minutes for $3000$ iterations to optimize a \threeD scene with accumulated HSIs constraints. 
In comparison, HolisticMesh \cite{weng2020holistic} which jointly optimizes human and a \threeD scene for one single image, directly trains the parameters of the network in Total3D~\cite{nie2020total3dunderstanding} to regress the \threeD scene, which is time-consuming and costs around 40 minutes.
For the human optimization, it runs twice in 5 minutes, i.e., the first pass is a HPS initialization used to refine the scenes, and the second pass is done using the refined scenes. In total, HolisticMesh takes 45 minutes for one single image. 
Our method takes almost the same time for a scene (around 10 objects) regardless how many frames in the input video. The number of frames in a video only influences the time of calculating the depth map, the SDF volume and the contact information of each body. However, this can be done once and is easily processed in parallel before the optimization.
In contrast, HolisticMesh \cite{weng2020holistic} processes a video sequentially, i.e., one frame after another. Therefore, the optimization time increases w.r.t.~the number of frames in a video.

\section{Sensitivity Analysis.}
Our approach uses HSIs observed in a video.
A longer video potentially has more HSIs, which results in more constraints for our objective function.
In \tref{tab:abl_len}, we analyze how different 
video lengths
influence 
scene reconstruction, 
by reporting the \threeD intersection-over-union (IoU) metric.
Specifically, we use $10$ sequences of the PROX qualitative dataset 
(one sequence per scene) 
and randomly sample $10$ segments of 
$10$s, $20$s, $30$s 
length from each sequence. 
We observe that longer 
sequences 
result in better performance, i.e., higher IoU 
and lower 
standard deviation.
We observe that
the performance of 3D scene reconstruction
depends 
on the \emph{number of HSIs} and not the video length, i.e., a short video with many HSIs 
results in 
a better reconstruction
than a long video with a few unique HSIs.

%

\begin{table}[h]
    \centering
    \small
    \begin{tabular}{c|ccc|c}
    
    & 10s & 20s & 30s & entire videos (51s)\\ \hline
   3D IoU mean $\uparrow$  & 0.389 & 0.395 & 0.407 & 0.424 \\
   3D IoU std. $\downarrow$ & 0.018 & 0.015 & 0.010 & - \\

    \end{tabular}
    \caption{Ablation study on different length of videos as input.
    The average length of entire videos is 51s.}
    
    \label{tab:abl_len}
\end{table}

We also do a sensitivity study w.r.t.~noise in the initialization.
In \tref{tab:noise}, we add uniform noise on the initial scale, translation and orientation of objects predicted by Total3D~\cite{nie2020total3dunderstanding}, and 
report the 
3D IoU.
%
MOVER is robust to noisy orientation and translation estimates from Total3D~\cite{nie2020total3dunderstanding}, but sensitive to the scale variation. 
%
This is because we 
currently regularize the optimization to the initial scale relatively strongly;
i.e., we cannot deviate much from a noisy estimate to ``correct'' it.
Relaxing 
$\mathcal{L}_\text{scale}$ easily resolves this.
%
%
\begin{table}[t]
    \centering
    \small
    \begin{tabular}{c|ccc}
    scale noise & $\pm$ 25\% & $\pm$ 15\% & $\pm$ 0.05\%  \\ 
    3D IoU $\uparrow$ & 0.345	& 0.3805	& 0.4105\\ \hline
    transl. & $\pm$ 30cm & $\pm$ 20cm & $\pm$10m  \\
    3D IoU $\uparrow$ & 0.4175	&0.416	&0.415 \\ \hline
    orien. & $\pm$45\degree & $\pm$30\degree & $\pm$15\degree  \\
    3D IoU $\uparrow$ & 0.4205	&0.418	&0.4205\\
    
    
    \end{tabular}
    \caption{Sensitivity analysis on scene reconstr.~with uniform noise on input scale, translation and orientation from Total3D~\cite{nie2020total3dunderstanding} (\textsl{Werkraum\_03301\_01} video). 
    Scene w/o noise has $0.417$ 3D IoU.
    }
    
    \label{tab:noise}
\end{table}

\section{More Evaluation Results on PROX Quantitative Dataset.}

\begin{table}[t]
\centering
\resizebox{\linewidth}{!}{
\begin{tabular}{cccccc}
    \toprule
     Methods & \multicolumn{3}{c}{Scene Recon.} & \multicolumn{2}{c}{HSI} \\ 
     \hline
      & $\text{IoU}_\text{3D}$ $\uparrow$ & P2S$\downarrow$ &  $\text{IoU}_\text{2D}$ $\uparrow$ & Non-Col $\uparrow$ & Cont. $\uparrow$ \\ \hline 
     HolisticMesh~\cite{weng2020holistic} & 0.239 & 0.133 & 0.533  & 0.948 & 0.951 \\
     Total3D~\cite{nie2020total3dunderstanding} & 0.063   & 0.409  & 0.342  & {0.940} &{0.436} \\   
     \rowcolor{lightgray}
    \textbf{Ours} & \textbf{0.390}  & \textbf{0.095} & \textbf{0.862} & \textbf{0.972} & \textbf{0.934} \\ 
    \bottomrule
    
\end{tabular}
}
\caption{Quantitative results for \threeD scene understanding (\threeD object detection) and human-scene interaction on the  \prox \textsl{quantitative} dataset. P2S, Non-Col and Cont denote \textsl{point2surface distance}, Non-Collision and Contactness respectively. }
\label{tab:quantitative_scene}
\end{table}

We also evaluate 3D scene reconstruction and human-scene interaction on PROX \textsl{quantitative}, as shown in \tref{tab:quantitative_scene}. Our method improves our input baseline \cite{nie2020total3dunderstanding} significantly and outperforms the previous method \cite{weng2020holistic} with a big margin in both 3D scene reconstruction metrics and human-scene interaction metrics.

\begin{figure} [t]
\centering
    \begin{tabular}{ccc}
        \hspace{-0.04\columnwidth} \includegraphics[width=0.33\columnwidth]{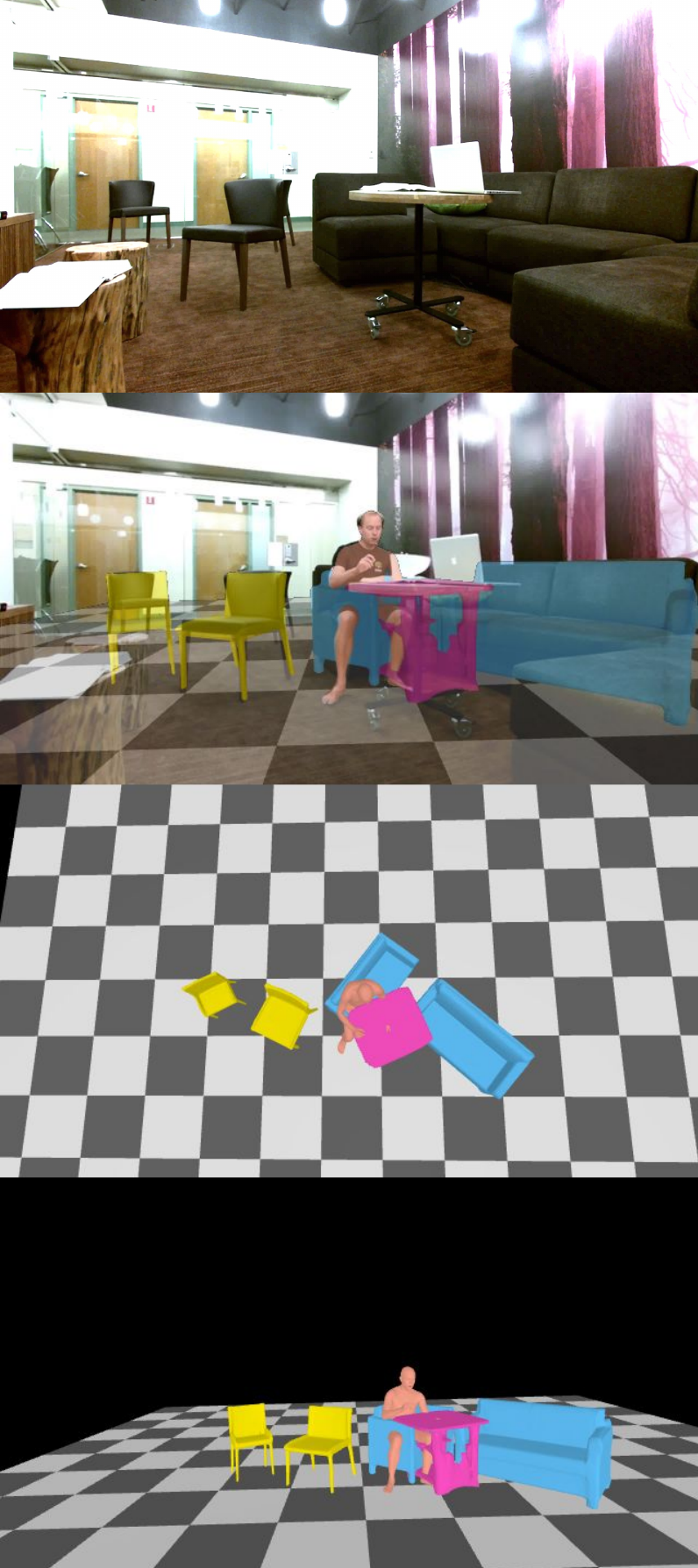}  &
         \hspace{-0.035\columnwidth}\includegraphics[width=0.33\columnwidth]{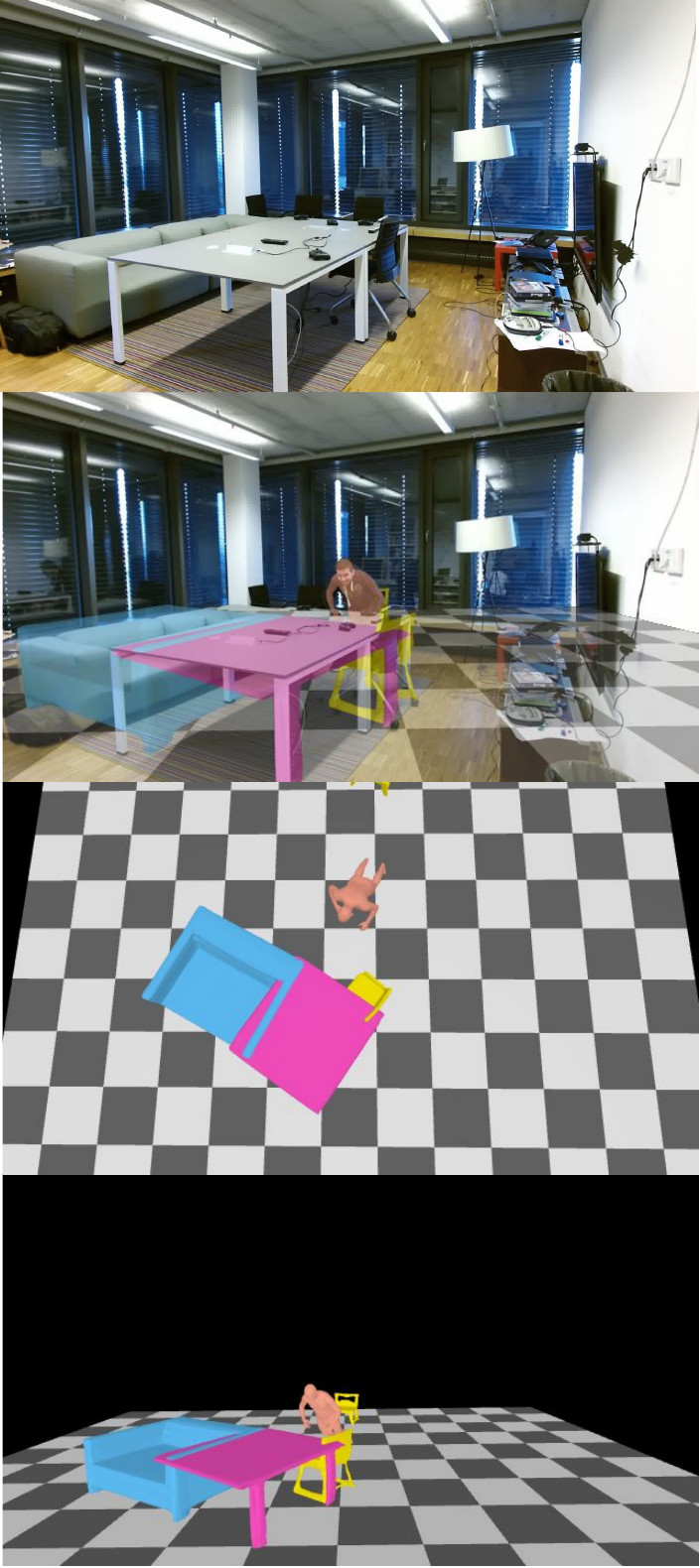} & 
         \hspace{-0.045\columnwidth} \includegraphics[width=0.33\columnwidth]{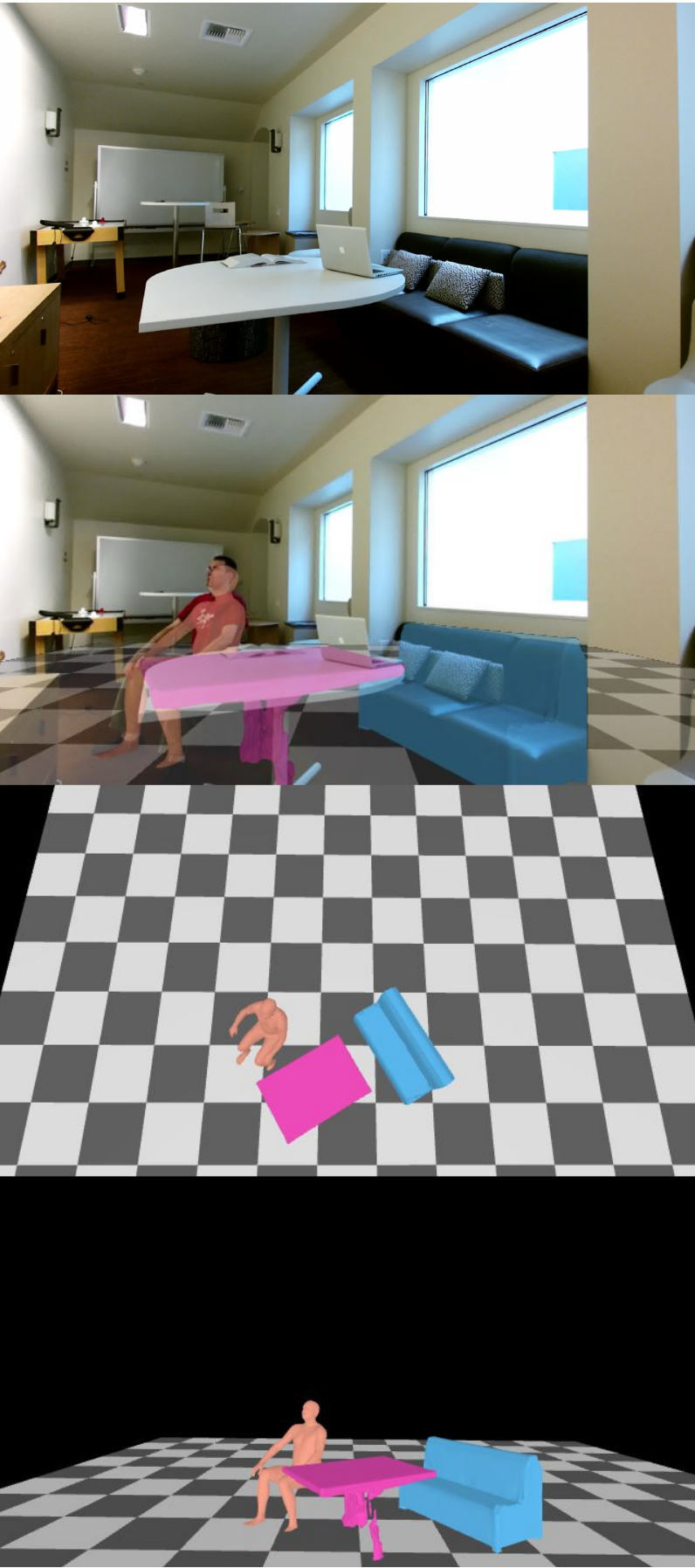}  
        \\
     (A) & (B) & (C) \\
    
    \end{tabular}
  \caption{Failure cases. (A) The estimated sofa has arms, which does not match the unarmed sofa in the input image. (B) The half bottom body is occluded, that leads to a wrong pose estimation as well as HSI observation. (C) The body is sitting ``in the air'', where the chair is missing.}
  \label{fig:failure_cases}
\end{figure}
\section{Failure Cases}
In this section, we discuss and show the failure cases of our method.
Besides optimizing the \threeD scene layout, we do not change the initial shape estimate of an object. Thus, wrong estimated geometry shape can still violate human's interaction, as shown in (A) in \fref{fig:failure_cases}.
A more flexible and adjustable geometry representation, e.g., an implicit representation, would be needed.
Human motion reconstruction struggles with severe occlusions in the input, that leads to wrong body poses as well as poor estimations of {\hsi}s, and, thus, influences our \threeD scene layout prediction, see (B) in \fref{fig:failure_cases}.
While not the scope of our work, the robustness and accuracy of human motion estimation can be improved by incorporating human motion priors or learning-based probabilistic human pose and estimation network.
Severe occlusion can also cause missing objects in the scene, like the chair in \fref{fig:failure_cases}(C).

In our pipeline, we currently consider the contact between \textsl{detected} objects and bodies. 
As a potential future extension of our method, one can also leverage the information from \twoD learning-based human-object interaction (HOI) detection network \cite{zou2021end}, by using contacted bodies to discover missing objects; or learn a model that jointly regress human-object interaction and their geometry shape.
\section{Additional Qualitative Results}
In \fref{fig:qualitative_1} and \fref{fig:qualitative_2}, we present additional qualitative results on PROX \cite{hassan2019resolving} qualitative and PiGraphs \cite{savva2016pigraphs} respectively.
As can be seen, our method performs well on a variety of different scenes and predicts a physically plausible scene layout.
We also refer to the suppl. video for results. 

\begin{figure*}[t]
\includegraphics[width=\textwidth]{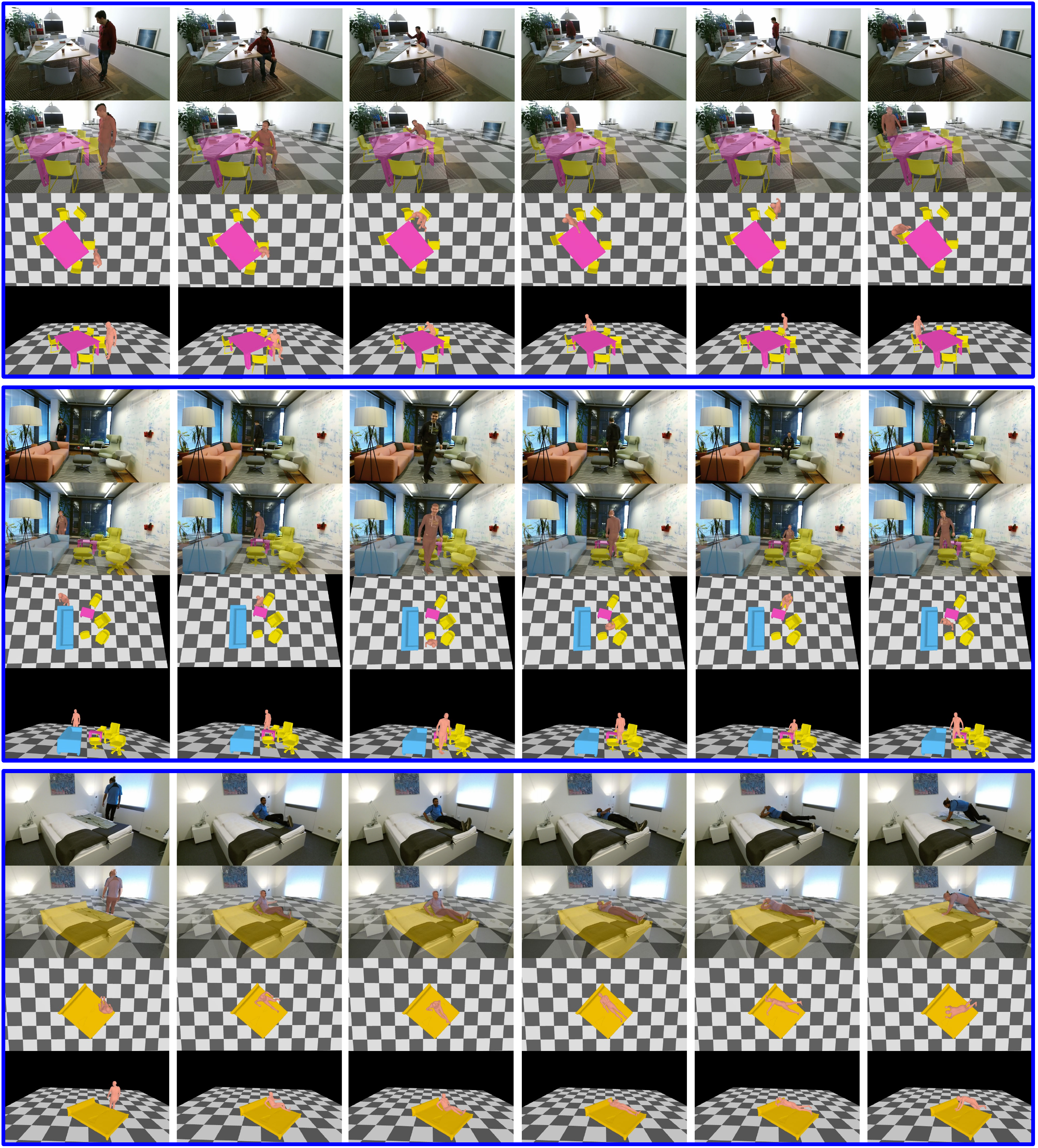}
\caption{More qualitative results on PROX qualitative dataset.}
\label{fig:qualitative_1}

\end{figure*}

\begin{figure*}[t]
\includegraphics[width=\textwidth]{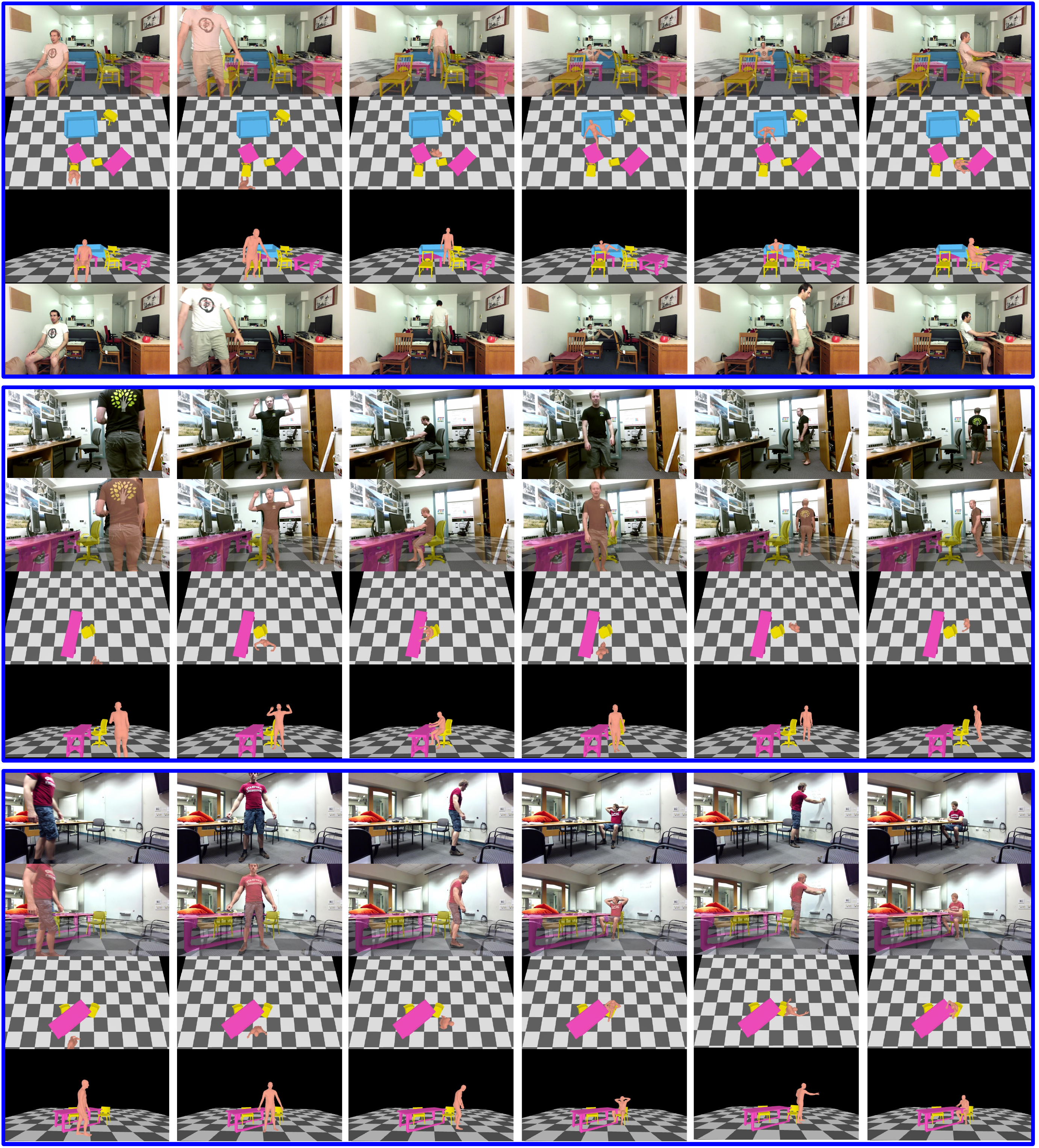}
\caption{More qualitative results on PiGraphs dataset.}
\label{fig:qualitative_2}

\end{figure*}

\section{Discussion of Potential Misuse}
Our approach is not intended for any surveillance application. Our goal is to understand how humans interact and move in scenes from videos (e.g., from TV sitcoms, movies, etc.), to this end both the scene geometry and the human pose and shape need to be reconstructed. 
Our method could be misused in potential surveillance applications that curtail human rights and
civil liberties, but we will restrict the usage of our method in a legal way.
\end{appendices}

\end{document}